\documentclass{article} 
\usepackage{iclr2022_conference,times}

\usepackage{hyperref}
\usepackage{url}
\usepackage[utf8]{inputenc} 
\usepackage[T1]{fontenc}    
\usepackage{hyperref}       
\usepackage{url}            
\usepackage{booktabs}       
\usepackage{amsfonts}       
\usepackage{nicefrac}       
\usepackage{microtype}      
\usepackage{xcolor}         
\usepackage{graphicx}
\usepackage{subfigure}
\usepackage[bb=boondox]{mathalfa}
\usepackage{amsmath}
\usepackage{wrapfig}
\usepackage{multirow}
\usepackage{algorithmic}
\usepackage{algorithm}
\usepackage{caption}
\usepackage[figuresright]{rotating}
\usepackage{chngpage}
\newcommand{\pluseq}{\mathrel{+}=}

\newcommand{\tabincell}[2]{\begin{tabular}{@{}#1@{}}#2\end{tabular}}
\usepackage[capitalize]{cleveref}

\newcommand{\name}{{MAANS}}
\newcommand{\planner}{{MSP}}

\title{Learning Efficient Multi-Agent Cooperative Visual Exploration}

\author{Chao Yu$^{1\sharp}$,
Xinyi Yang$^{1*}$,  
Jiaxuan Gao$^{2*}$, 
Huazhong Yang$^{1}$,
Yu Wang$^{1}$, 
Yi Wu$^{23\natural}$
\\
$^1$ Department of Electronic Engineering, Tsinghua University\\ [0.5ex]
$^2$ Institute for Interdisciplinary Information Sciences, Tsinghua University\\ [0.5ex]
$^3$ Shanghai Qi Zhi Institute\\ [0.5ex]
$^{\sharp}$\texttt{zoeyuchao@gmail.com}, $^{\natural}$\texttt{jxwuyi@gmail.com} \\
}

\iclrfinalcopy 
\begin{document}
\maketitle

\begin{abstract}
We tackle the problem of cooperative visual exploration where multiple agents need to jointly explore unseen regions as fast as possible based on visual signals. Classical planning-based methods often suffer from expensive computation overhead at each step and a limited expressiveness of complex cooperation strategy. By contrast, reinforcement learning (RL) has recently become a popular paradigm for tackling this challenge due to its modeling capability of arbitrarily complex strategies and minimal inference overhead. In this paper, we extend the state-of-the-art single-agent visual navigation method, \emph{Active Neural SLAM} (ANS), to the multi-agent setting by introducing a novel RL-based planning module, \emph{\underline{M}ulti-agent \underline{S}patial \underline{P}lanner} (\planner).
{\planner} leverages a transformer-based architecture, \emph{Spatial-TeamFormer}, which effectively captures spatial relations and intra-agent interactions via hierarchical spatial self-attentions. In addition, we also implement a few multi-agent enhancements to process local information from each agent for an aligned spatial representation and more precise planning. 
Finally, we perform policy distillation to extract a meta policy to significantly improve the generalization capability of final policy. 
We call this overall solution, \emph{\underline{M}ulti-\underline{A}gent \underline{A}ctive \underline{N}eural \underline{S}LAM} (\name).  {\name} substantially outperforms classical planning-based baselines for the first time in a photo-realistic 3D simulator, Habitat. 
Code and videos can be found at {\footnotesize{\url{https://sites.google.com/view/maans}}}.
\end{abstract}

\section{Introduction}
\label{sec:intro}

Visual exploration~\citep{ramakrishnan2021exploration} is an important task for building intelligent embodied agents and has been served as a fundamental building block for a wide range of applications, such as scene reconstruction~\citep{anguelov2010google,isler2016information}, autonomous driving~\citep{autonomousdriving}, disaster rescue~\citep{rescue} and planetary exploration~\citep{DBLP:journals/corr/abs-2002-00515}. In this paper, we consider a multi-agent exploration problem, where multiple homogeneous robots simultaneously explore an unknown spatial region via visual and sensory signals in a cooperative fashion. The existence of multiple agents enables complex cooperation strategies to effectively distribute the workload among different agents, which could lead to remarkably higher exploration efficiency than the single-agent counterpart. 

Planning-based solutions have been widely adopted for robotic navigation problems for both single-agent and multi-agent scenarios~\citep{frontier3,singleagent-RL2,RRT}. Planning-based methods require little training and can be directly applied to different scenarios. However, these methods often suffer from limited expressiveness capability on coordination strategies, require non-trivial hyper-parameter tuning for each test scenario, and are particularly time-consuming due to repeated re-planning at each decision step. 
By contrast, reinforcement learning (RL) has been promising solution for a wide range of decision-making problems~\citep{RL1,RL2}, including various visual navigation tasks~\citep{ans,singleagent-RL1,singleagent-RL2}. An RL-based agent is often parameterized as a deep neural network and directly produces actions based on raw sensory signals. Once a policy is well trained by an RL algorithm, the robot can capture arbitrarily complex strategies and produce real-time decisions with efficient inference computation (i.e., a single forward-pass of neural network). 



However, training effective RL policies can be particularly challenging. 
This includes two folds: (1) learning a cooperative strategy over multiple agents in an end-to-end manner becomes substantially harder thanks to an exponentially large action space and observation space when tackling the exploration task based on visual signals; (2) RL policies often suffer from poor generalization ability to different scenarios or team sizes compared with classical planning-based approaches. Hence, most RL-based visual exploration methods focus on the single-agent case~\citep{ans,singleagent-RL1,singleagent-RL2} or only consider a relatively simplified multi-agent setting (like maze or grid world~\citep{multiagent-RL}) of a fixed number of agents~\citep{liu2021multi}.

In this work, we develop \emph{\underline{M}ulti-\underline{A}gent \underline{A}ctive \underline{N}eural \underline{S}LAM} (\name), the first RL-based solution for cooperative multi-agent exploration that substantially outperforms classical planning-based methods in a photo-realistic physical simulator, Habitat~\citep{habitat}. {\name} extends the single-agent Active Neural SLAM method~\citep{singleagent-RL2} to the multi-agent setting. In {\name}, an agent consists of 4 components, a neural SLAM module, a planning-based local planner, a local policy for control, and the most critical one, 
a novel \emph{\underline{M}ulti-agent \underline{S}patial \underline{P}lanner (\planner)}. which is an RL-trained planning module that can capture complex intra-agent interactions via a self-attention-based architecture, \emph{Spatial-TeamFormer}, and produce effective navigation targets for a varying number of agents.
We also implement a map refiner to align the spatial representation of each agent's local map, and a map merger, which enables the local planner to perform more precise sub-goal generation over a manually combined approximate 2D map. 
Finally, instead of directly running multi-task RL  over all the training scenes, we first train a single policy on each individual scene and then use policy distillation to extract a meta policy, leading to a much improved generalization capability,

We conduct thorough experiments in a photo-realistic physical simulator, Habitat, and compare {\name} with a collection of classical planning-based methods and RL-based variants. Empirical results show that {\name} has a 20.56\% and 7.99\% higher exploration efficiency on training and testing scenes 
than the best planning-based competitor. The learned policy can further generalize to novel team sizes in a zero-shot manner as well.

\section{Related Work}
\label{sec:related}

\subsection{Visual Exploration}
In classical visual exploration solutions, an agent first locates its position and re-constructs the 2D map based on its sensory signals, which is formulated as Simultaneous Localization and Mapping (SLAM)~\citep{SLAM1}. Then a search-based planning algorithm will be adopted to generate valid exploration trajectories. 
Representative variants include frontier-based methods~\citep{RRT,frontier1,APF}, which always choose navigation targets from the explored region, and sampling-based methods~\citep{sample1}, which generate goals via a stochastic process. 
In addition to the expensive search computation for planning, these methods do not involve learning and thus have limited representation capabilities for particularly challenging scenarios. 
Hence, RL-based methods have been increasingly popular for their training flexibility and strong expressiveness power. Early methods simply train navigation policies in a purely end-to-end fashion~\citep{singleagent-RL1,jain2019two} while recent works start to incorporate the inductive bias of a spatial map structure into policy  representation by introducing a differentiable spatial memory~\citep{neuralmap1,semantic-RL1,neuralmap2}, semantic prior knowledge~\citep{liu2021multi} or learning a topological scene graph~\citep{semanticmap1,singleagent-RL4,semantic-RL2}. 


The Active Neural SLAM (ANS) method~\citep{ans} is the state-of-the-art framework for single-agent visual exploration, which takes advantage of both planning-based and RL-based techniques via a modular design (details in Sec.~\ref{ANS}). 
There are also follow-up enhancements based on the ANS framework, such as improving map reconstruction with occupancy anticipation~\citep{singleagent-RL9} and incorporating semantic signals into the reconstructed map for semantic exploration~\citep{semanticmap3}.
Our {\name} can be viewed as a multi-agent extension of ANS with a few multi-agent-specific components.



\subsection{Multi-agent Cooperative Exploration} There have been works extending planning-based visual exploration solutions to the multi-agent setting by introducing handcraft planning heuristics over a shared reconstructed 2D map~\citep{vcap2013multi,multi-classical2,desaraju2011decentralized,Voronoi,WMA-RRT,patelmulti,multi-classical1}. However, due to the lack of learning, these methods may have the limited potential of capturing non-trivial multi-agent interactions in challenging domains. 
By contrast, multi-agent reinforcement learning (MARL) has shown its strong performances in a wide range of domains~\citep{MARL-review1}, so many works have been adopting MARL to solve challenging cooperative problems. Representative works include value decomposition for approximating credit assignment~\citep{qmix,vdn}, learning intrinsic rewards to tackle sparse rewards~\citep{iqbal2019coordinated,liu2021cooperative,Wang*2020Influence-Based} and curriculum learning~\citep{epciclr2020,wang2020few}.

However, jointly optimizing multiple policies makes multi-agent RL training remarkably more challenging than its single-agent counterpart. Hence, these end-to-end RL methods either focus on much simplified domains, like grid world or particle world~\citep{multiagent-RL}, or still produce poor exploration efficiency compared with classical planning-based solutions. 
Our {\name} framework adopts a modular design and is the first RL-based solution that significantly outperforms classical planning-based baselines in a photo-realistic physical environment.


Finally, we remark that {\name} utilizes a centralized global planner {\planner}, which assumes perfect communication between agents. There are also works on multi-agent cooperation with limited or constrained communication~\citep{jiang2018learning,DBLP:journals/corr/PengYWYTLW17,sukhbaatar2016learning,foerster2016learning,jain2019two,multiagent-navigation1,multiagent-RL1}, which are parallel to our focus.

\subsection{Size-Invariant Representation Learning}

There has been rich literature in deep learning studying representation learning over an arbitrary number of input entities in deep learning~\citep{graphnetwork,deepset}.
In MARL, the self-attention mechanism~\citep{transformer} has been the most popular policy architecture to tackle varying input sizes~\citep{duan2017one,dgn,hama,wang2018non} or capture high-order relations~\citep{iqbal2019actor,malysheva2018deep,semantic-RL2,zambaldi2018relational}. A concurrent work~\citep{multiagent-navigation1} also considers the zero-shot team-size adaptation in the photo-realistic environment by learning a simple attention-based communication channel between agents. By contrast, our works develop a much expressive network architecture, Spatial-TeamFormer, which adopts a hierarchical self-attention-based architecture to capture both intra-agent and spatial relationships and results in substantially better empirical performance (see~\cref{sec: planner}).
Besides, parameter sharing is another commonly used technique in MARL for team-size generalization, which has been also shown to help reduce nonstationarity and accelerate training~\citep{DBLP:journals/corr/abs-1710-00336,DBLP:journals/corr/abs-2005-13625}. Our work follows this paradigm as well.

\section{Preliminary}
\label{sec:preliminary}

\subsection{Task Setup}

We consider a multi-agent coordination indoor active SLAM problem, in which a team of agents needs to cooperatively explore an unknown indoor scene as fast as possible. 
At each timestep, each agent performs an action among \emph{Turn Left}, \emph{Turn Right} and \emph{Forward}, and then receives an RGB image through a camera and noised pose change through a sensor, which is provided from the Habitat environment.
We consider a decision-making setting by assuming perfect communication between agents.
The objective of the task is to maximize the accumulative explored area within a limited time horizon.


\subsection{Active Neural SLAM}

\label{ANS}
The ANS framework~\citep{ans} consists of 4 parts: a neural SLAM module, a RL-based global planner, a planning-based local planner and a local policy. 
The neural SLAM module, which is trained by supervised learning, takes an RGB image, the pose sensory signals, and its past outputs as inputs, and outputs an updated 2D reconstructed map and a current pose estimation. Note that in ANS, the output 2D map only covers a neighboring region of the agent location and always keeps the agent at the egocentric position. For clarification, we call this raw output map from the SLAM module a \emph{agent-centric local map}. 

The global planner in ANS takes in an augmented agent-centric \emph{local map}, which includes channels indicating explored regions, unexplored regions and obstacles and the history trajectory, as its input, and outputs two real numbers from two Gaussian distributions denoting the coordinate of the long-term goal. This global planner is parameterized as a CNN policy and trained by the PPO algorithm~\citep{ppo}.
The local planner performs classical planning, i.e., Fast Marching Method (FMM)~\citep{fmm}, over the agent-centric local map towards a given long-term goal, and outputs a trajectory of short-term sub-goals.
Finally, the local policy produces actions given an RGB image and a sub-goal and is trained by imitation learning.


\begin{figure*}[h]
	\centering
	
    \includegraphics[width=0.9\linewidth]{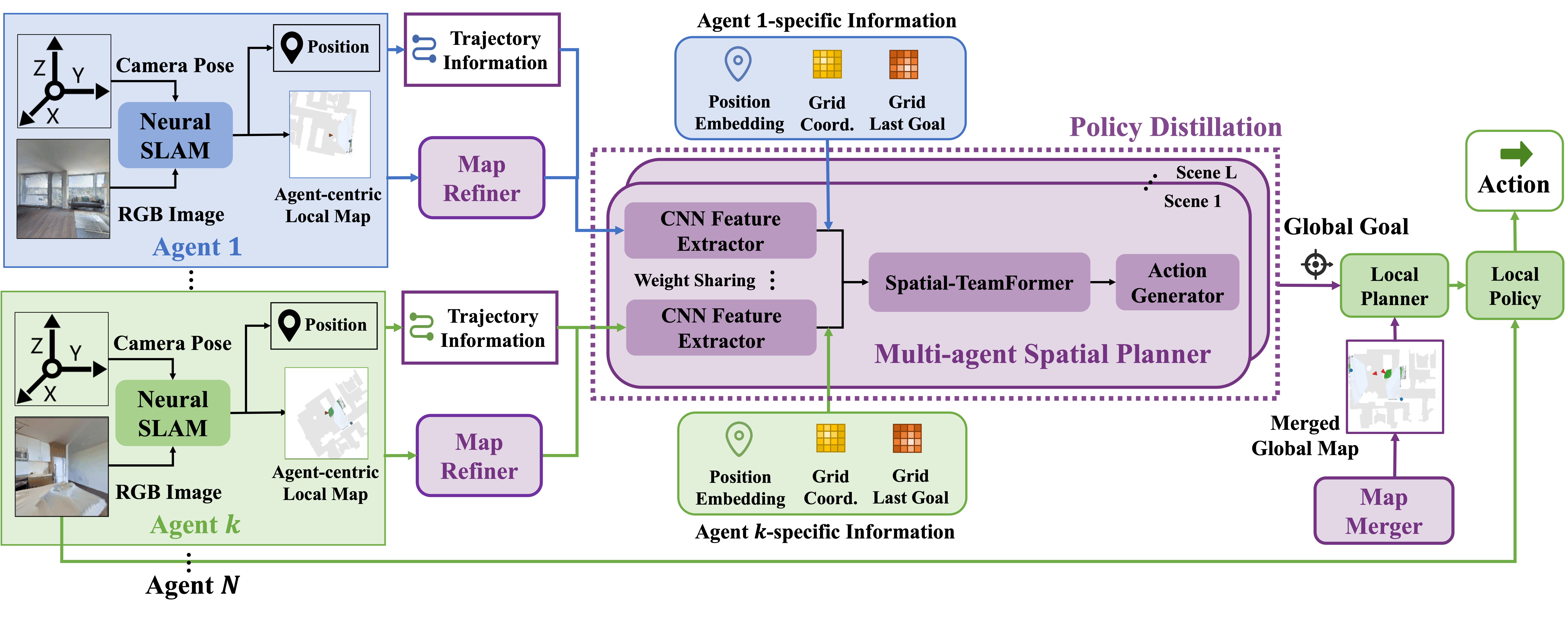}
    
	\centering \caption{Overview of \emph{Multi-Agent Active Neural SLAM} (\name). }
\label{fig:Overview}

\end{figure*}
With the advantage of the modeling capability of arbitrarily complex strategy in RL, an RL-based global planner which determines the global goals encourages exploration faster. To apply RL training, we model the problem as a decentralized partially observable Markov decision process (Dec-POMDP). Dec-POMDP is parameterized by $\langle S, A, O, R, P, n, \gamma, h\rangle$. $n$ is the number of agents. $S$ is the state space, $A$ is the joint action space. $o^{(i)}=O(s;i)$ is agent $i$'s observations at state $s$. $P(s'|s, a)$ defines the transition probability from state $s$ to state $s'$ via joint action $a$. $R(s,A)$ is the shared reward function. $\gamma$ is the discount factor. The objective function is $J(\theta)=\mathbb E_{a, s}[\sum_t\gamma^tR(s^t, a^t)]$. In this task, the policy $\pi_\theta$ generates a global goal for each agent every decision-making step. The shared reward function is defined as the accumulative environment reward every global goal planning step.

\section{Methodology}
\label{sec:method}



The overview of {\name} is demonstrated in ~\cref{fig:Overview}, where each agent is presented in a modular structure.
When each agent receives the visual and pose sensory signals from the environment, the \emph{Neural SLAM} module corrects the sensor error and performs SLAM in order to build a top-down 2D occupancy map that includes explored area and discovered obstacles.
Then we use a \emph{Map Refiner} to rotate each agent's egocentric local map to a global coordinate system. We augment these refined maps with each agent's trajectory information and feed these spatial inputs along with other agent-specific information to our core planning module, \emph{\underline{M}ulti-agent \underline{S}patial \underline{P}lanner (\planner)} to generate a global goal as the long-term navigation target for each individual agent. We remark that only estimated geometric information is utilized in this map fusion process.
To effectively reach a global goal, the agent first plans a path to this long-term goal in a manually merged global map using FMM and generates a sequence of short-term sub-goals. Finally, given a short-term sub-goal, a \emph{Local Policy} outputs the final navigation action based on the visual input and the relative spatial distance as well as the relative angle to the sub-goal. 

Note that the Neural SLAM module and the Local Policy do not involve multi-agent interactions, so we directly reuse these two modules from ANS~\citep{singleagent-RL2}.
We fix these modules throughout training and only train the planning module {\planner} using the MAPPO algorithm~\citep{mappo}, a multi-agent variant of PPO~\citep{ppo}. Hence, the actual action space for training {\planner} is the spatial location of the global goal. 



\subsection{Multi-agent Spatial Planner}

\begin{figure*}[h]
	\centering

    \includegraphics[width=0.8\linewidth]{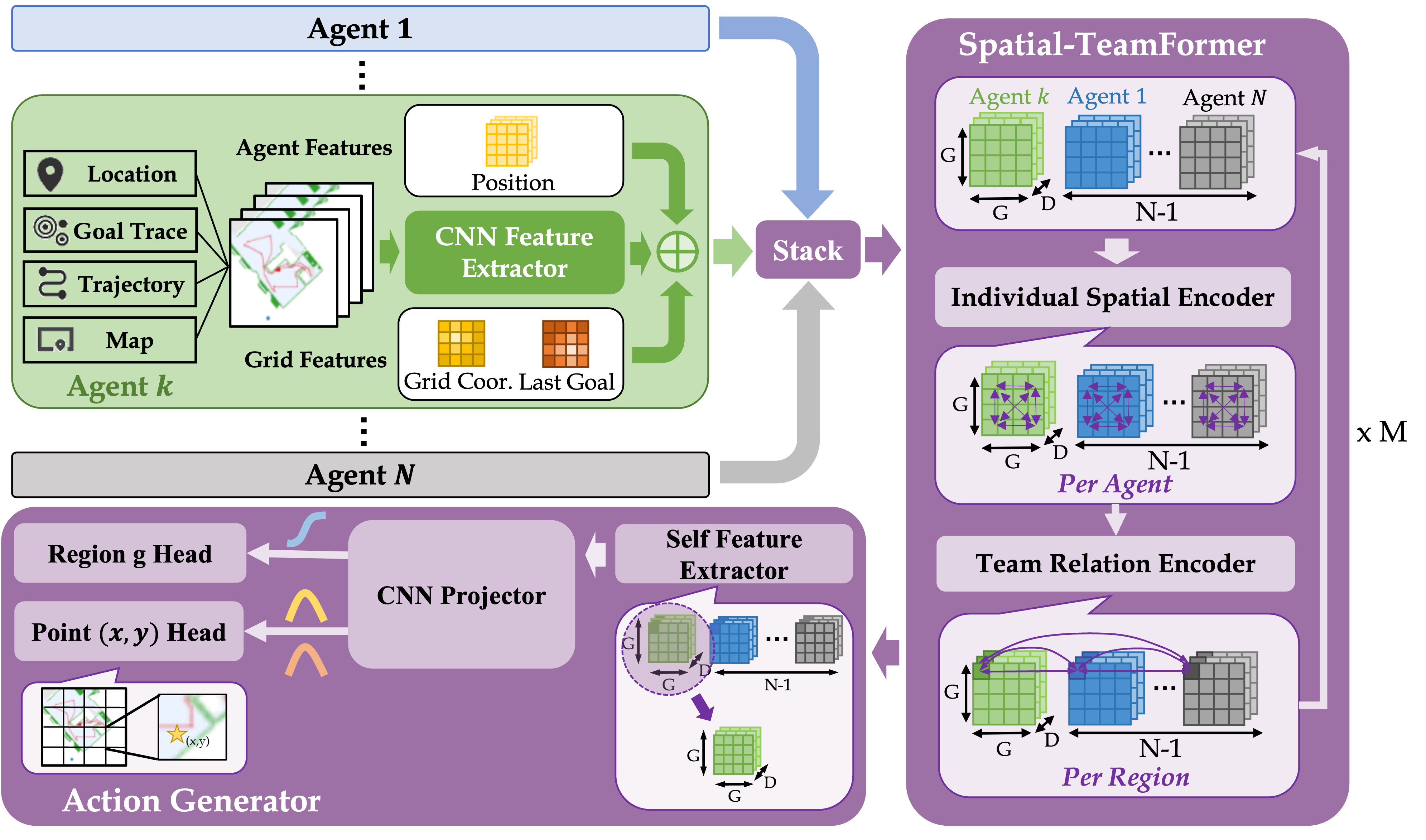}
 
	\centering \caption{Workflow of \emph{Multi-agent Spatial Planner} (\planner), including a CNN-based Feature Extractor, a Spatial-TeamFormer for representation learning and an Action Generator.}
\label{fig:acp}

\end{figure*}

Multi-agent Spatial Planner (\planner) is the core component in {\name}, which could perform planning for an arbitrary number of agents. The full workflow of {\planner} is shown in~\cref{fig:acp}. 
{\planner} first applies a weight-shared CNN feature extractor to extract spatial feature maps from each agent's local navigation trajectory and then fuses team-wise information with hierarchical transformer-based network architecture, Spatial-TeamFormer. Finally, an action generator will generate a spatial global goal based on the features from Spatial-TeamFormer. 
Suppose there are a total of $N$ agents and the current decision-making agent has ID $k$. We will describe how agent $k$ generates its long-term goal via the 3 parts in {\planner} in the following content. Note that due to space constraints, we only present the main ideas while more computation details can be found in Appendix \ref{app: msp}. 


\subsubsection{CNN Feature Extractor}

For every single agent, we use its current location, movement trajectory, previous goal, goal history, self-occupancy map and obstacle map as inputs and convert them to a $480\times 480$ 2D map with 6 channels over a global coordinate system. 
We adopt a weight-shared CNN network with 5 layers to process each agent's input map, which produces a $G\times G$ feature map with $D=32$ channels. $G$ corresponds to the discretization level of the scene. We choose $G=8$ in our work, leading to $G^2$ grids corresponding to different spatial regions in the underlying indoor scene. 

Besides CNN spatial maps, we also introduce additional features, including agent-specific embeddings of its current position and grid features, i.e., the embeddings of the relative coordinate of each grid to the agent position as well as the embedding of the previous global goal. 





\subsubsection{Spatial-TeamFormer}
With a total of $N$ extracted $G\times G$ feature maps, we aim to learn a team-size-invariant spatial representation over all the agents.
Transformer has been a particularly popular candidate for learning invariant representations, but it may not be trivially applied in this case.
Standard Transformer model in NLP~\citep{transformer} tackles 1-dimensional text inputs, which ignores the spatial structure of input features. Visual transformers~\citep{vit} capture spatial relations well by performing spatial self-attention. However, we have a total of $N$ spatial inputs from the entire team. 

Hence, we present a specialized architecture to jointly leverage intra-agent and spatial relationships in a hierarchical manner, which we call \emph{Spatial-TeamFormer}.
A Spatial-Teamformer block consists of two layers, i.e., an \emph{Individual Spatial Encoder} for capturing spatial features for each agent, and a \emph{Team Relation Encoder} for reasoning cross  agents. 
Similar to visual transformer~\citep{vit}, \emph{Individual Spatial Encoder} focuses only on spatial information by performing a spatial self-attention over each agent's own $G\times G$ spatial map without any cross-agent computation. 
By contrast,  \emph{Team Relation Encoder} completely focuses on capturing team-wise interactions without leveraging any spatial information. In particular, for each of the $G\times G$ grid, \emph{Team Relation Encoder} extracts the features w.r.t. that grid from the $N$ agents and performs a standard transformer over these $N$ features.
We can further stack multiple Spatial-TeamFormer blocks for even richer interaction representations. 

We remark that another possible alternative to Spatial-TeamFormer is to simply use a big transformer over the aggregated $N\times G\times G$ features. Such a naive solution is substantially more expensive to compute ($O(N^2G^4)$ time complexity) than Spatial-TeamFormer ($O(N^2G^2+NG^4)$ time complexity), which may also incur significant learning difficulty in practice (see~\cref{sec: ST}).

\subsubsection{Action Generator}

\label{sec:spatial-action-decoder}

The Action Generator is the final part of {\planner}, which outputs a long-term global goal over the reconstructed map.
Since spatial-TeamFormer produces a total of $N$ rich spatial representation, which can be denoted as $N \times G \times G$ , we take the first $G\times G$ grid, which is the feature map of the current agent, to derive a single team-size-invariant representation. 


In order to produce accurate global goals, we adopt a spatial action space with two separate action heads, i.e., a discrete region head for choosing a region $g$ from the $G\times G$ discretized grids, and a continuous point head for outputing a coordinate $(x,y)$, indicating the relative position of the global goal within the selected region $g$. To compute the action probability for $g$, we compute a spatial softmax operator over all the grids while to ensure the scale of $(x,y)$ is bounded between 0 and 1, we apply a sigmoid function before outputting the value of $(x,y)$.
We remark that such a spatial design of action space is beneficial since it alleviates the problem of multi-modal issue of modeling potential "good" goals, which could not be simply represented by a simple normal distribution as used in \citep{singleagent-RL1} (see \cref{sec: planner}).

\subsection{Map Refiner for Aligned 2D Maps}

\begin{figure}[ht!]

	\centering
	\vspace{3pt}
    \includegraphics[width=0.8\linewidth]{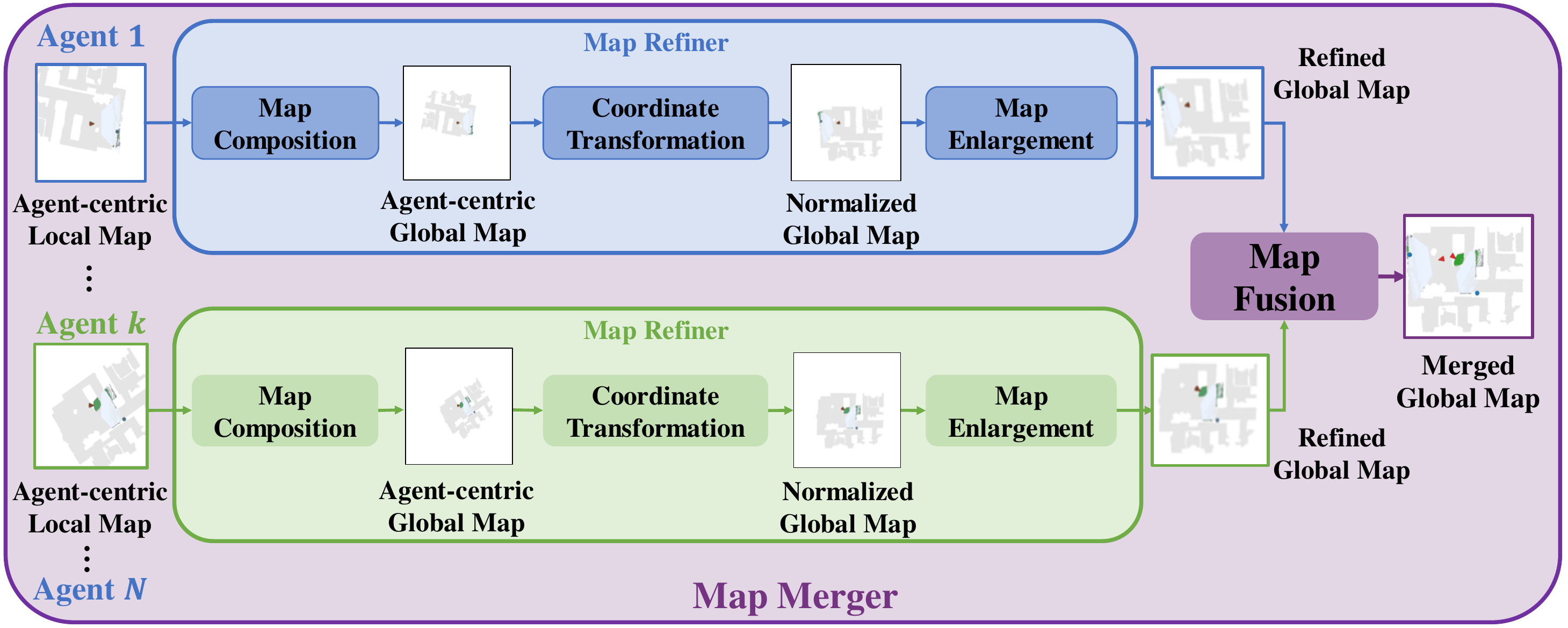}
	\centering

	\caption{Computation workflow of \emph{map refiner} (blue and green) and \emph{map merger} (purple).}
\label{fig:merger}

\end{figure}

We develop a map refiner to ensure all the maps from the neural SLAM module are within the same coordinate system. The workflow is shown as the blue and green part in Fig.~\ref{fig:merger}. The map refiner first composes all the past \emph{agent-centric local maps} to recover the agent-centric \emph{global} map. Then, we transform the coordinate system based on the pose estimates to normalize the global maps from all the agents w.r.t. the same coordinate system. 
Note that when an agent explores the border of the house, the agent-centric local map often covers a large portion of invisible region.
As a result, the normalized global map will accordingly contain a large unexplorable boundary surrounding the actual explorable house region. 
To ensure the feature extractor in {\planner} concentrates only on the viable part and also induce a more focused spatial action space, we crop the unexplorable boundary of the normalized map and enlarge the house region as our final refined map.

\subsection{Map Merger for Improved Local Planning}

The local planner from ANS plans sub-goals on the agent-centric local map, while in our setting, we can also leverage the information from other agents to plan over a more accurate map. The diagram of map merger is shown in Fig.~\ref{fig:merger}.
After obtaining $N$ enlarged global maps via the map refiner, the map merger simply integrates all these maps by applying a max-pooling operator for each pixel location. That is, for each pixel in the merged global map, the probability of it being an obstacle is the maximum value at that pixel over all individual enlarged global maps. We remark that the artificial merged global map is only utilized in the local planner, but not in the global planner {\planner}. We empirically observe that having a coarse merged map produces better short-term local goal while such an artificial map is not sufficient for accurate global planning. (see~\cref{sec: planner})

\subsection{Policy Distillation for Improved Generalization}

The common training paradigm for visual exploration is multi-task learning, i.e., at each training episode, a random training scene or team size is sampled and all collected samples are aggregated for policy optimization~\citep{ans,singleagent-RL1}.
However, we empirically observe that different Habitat scenes and team sizes may lead to drastically different exploration difficulties. 
During training, gradients from different configurations may negatively impact each other. Similar observations have been also reported in the existing literature~\citep{hessel2019multi,teh2017distral}. We use policy distillation to tackle this problem. 
Therefore, we adopt a two-phase distillation-based solution: in the first phase, we train separate policies for representative training scenes with a fixed team size, i.e., we choose $N=2$ in our experiments
in the second phase, we learn another policy with $N=2$ agents to distill the collection of pretrained policies over different training scenes and directly measure the generalization ability of this distillation policy to novel scenes and different team sizes.
More specifically, for the $i$-th training scene, we first learn a specialized teacher policy $\pi(g,x,y|s,\theta_i)$ given state $s$ with parameter $\theta_i$, where $g$ denotes the region output and $(x,y)$ is the point head output. 
Then we train another distillation policy $\pi(g,x,y|s,\theta)$ by simply running a dagger-style imitation learning, i.e., randomly rollout trajectories w.r.t. the distillation policy $\pi(s,\theta)$ and imitate the output from the specific teacher policy. 
Since the region action $g$ is discrete, we adopt a KL-divergence-based loss function while for the continuous point action $(x,y)$, a squared difference loss between the teacher policy and distillation policy is optimized. 



\section{Experiment Results}
\label{sec:exp}

\subsection{Experiment Setting}

We adopt scene data from the Gibson Challenge dataset~\citep{dataset} while the visual signals and dynamics are simulated by the Habitat simulator~\citep{habitat}. Although Gibson Challenge dataset provides $72$ training and $14$ validation scenes, we discard scenes that are not appropriate for our task, such as scenes that have large disconnected regions or multiple floors so that the agents are not possible to achieve 90\% coverage of the entire house.
Then we categorize the remaining scenes into $23$ training scenes
and $10$ testing scenes.
We consider $N=2,3,4$ agents in our experiments.
Every RL training is performed with $10^4$ training episodes over 3 random seeds. Each evaluation score is expressed in the format of ``mean (standard deviation)'', which is averaged over a total of 300 testing episodes, i.e., 100 episodes per random seed. More details are deferred to Appendix \ref{app: exp}.


\subsection{Evaluation Metrics}

We take 3 metrics to examine the exploration efficiency:
\begin{enumerate}
    \item \textbf{Coverage}: \emph{Coverage} represents the ratio of areas explored by the agents to the entire explorable space in the indoor scene at the end of the episode. Higher \emph{Coverage} implies more effective exploration. 
    \item \textbf{Steps}: \emph{Steps} is the number of timesteps used by agents to achieve a coverage ratio of $90\%$ within an episode. Fewer \emph{Steps} implies faster exploration.
    \item \textbf{Mutual Overlap}: For effective collaboration, each agent should visit regions different from those explored by its teammates. We report the average overlapping explored area over each pair of agents when the coverage ratio reaches 90\%. \emph{Mutual Overlap} denotes the normalized value of this metric. Lower \emph{Mutual Overlap} suggests better multi-agent coordination. 
\end{enumerate}
\vspace{-2mm}



\subsection{Baselines}

We first adapt $3$ single-agent planning-based methods, namely \emph{Nearest}~\citep{frontier1}, \emph{Utility}~\citep{Utility}, and \emph{RRT}~\citep{RRT}, to our problems by planning on the merged global map. The 3 planning-based baselines are frontier-based, i.e., they choose long-term navigation goals from the boundary between currently explored and unexplored area using different heuristics: \emph{Nearests} chooses the nearest candidate point; \emph{Utility} measures a hand-crafted utility function; \emph{RRT} develops a Rapid-exploring Random Tree and selects the best candidate from the tree through an iterative process. Note that though these are originally single-agent methods and are adapted to multi-agent settings by planning on the merged global map. When choosing global goals, each agent performs computation based on the merged global map, its current position and its past trajectory.

\begin{figure*}[ht!]
	\centering

    \includegraphics[width=0.9\linewidth]{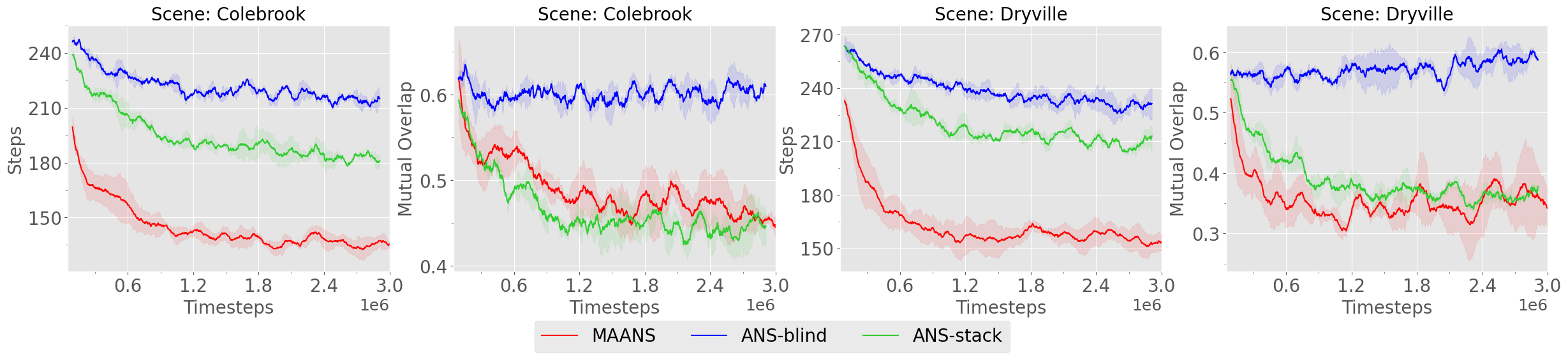}
    
	\centering 
	\caption{Comparison between {\name} (red) and other ANS variants.}
\label{fig:var}
\end{figure*}

For multi-agent baselines, we compare {\name} with $3$ planning-based methods, namely \emph{Voronoi}~\citep{Voronoi}, \emph{APF}~\citep{APF} and \emph{WMA-RRT}~\citep{WMA-RRT}.
\emph{APF}~\citep{APF} computes artificial potential field over clustered frontiers and plans a potential-descending path with maximum information gain. APF introduces resistance force among multiple agents to avoid repetitive exploration. \emph{WMA-RRT}~\citep{WMA-RRT} is a multi-agent variant of RRT, in which agents cooperatively maintain a single tree and follow a formal locking-and-search scheme. \emph{Voronoi}-based method partitions the map into different parts using a voronoi partition and each agent only searches unexplored area in its own partition. 

Finally, we also evaluate the performance of random policies for references. We remark that all the baselines only replace the global planner module with corresponding planning-based methods while utilizing the same neural SLAM, local planner and local policy modules as {\name} for a fair comparison. More implementation details can be found in Appendix \ref{app: baseline}.

\subsection{Ablation Study}

We report the training performances of multiple RL variants on 2 selected scenes, \textit{Colebrook} and \textit{Dryville}, and measure the \emph{Steps} and the \emph{Mutual Overlap} over these 2 scenes.

\subsubsection{Comparison with ANS variants}
We first consider 2 ANS variants, ANS-blind and ANS-stack, other than {\name}.

\begin{itemize}
\vspace{-1mm}
    \item \textbf{ANS-blind} We train $N$ ANS agents to explore blindly, i.e., without any communication, in the environment.
    \item \textbf{ANS-stack} We directly stack all the agent-centric local maps from the neural SLAM module as the input representation to the global planner, and retrain the ANS global planner under our multi-agent task setting.
\end{itemize}

We demonstrate the training curves in Fig.~\ref{fig:var}. Regarding the \emph{Steps}, both ANS variants perform consistently worse than {\name} on each map. Regarding the \emph{Mutual Overlap}, the blind variant fails to cooperate completely while the stack variant produces comparable \emph{Mutual Overlap} to {\name} despite its low exploration efficiency. We remark that ANS-stack performs global and local planning completely on the agent-centric \emph{local} map while the local map is a narrow sub-region over the entire house, which naturally leads to a much conservative exploration strategy and accordingly helps produces a lower \emph{Mutual Overlap}.






\begin{figure}[ht!]
	\centering

    \includegraphics[width=0.9\linewidth]{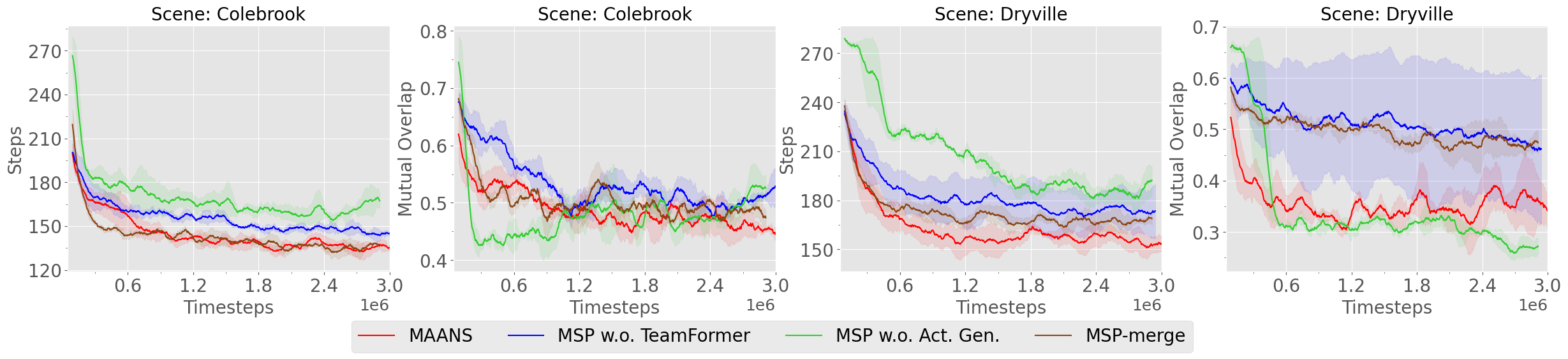}

	\centering 
	\caption{Ablation studies on {\planner} components.}
\label{fig:ab-scp}

\end{figure}

\subsubsection{Ablation Study on {\planner}}
\label{sec: planner}
We consider 3 additional {\planner} variants:

\begin{itemize}
    \item \textbf{{\planner} w.o. TeamFormer}: We completely substitute Spatial-TeamFormer with a simple average pooling layer over the extracted spatial features from CNN extractors.
    \item \textbf{{\planner} w.o. Act. Gen.} We remove the region head from the spatial action generator, so that the global goal is directly generated over the entire refined global map via two Gaussian action distributions. We remark that such an action space design follows the original ANS paper~\citep{singleagent-RL2}.
    \item \textbf{{\planner}-merge} We consider another {\planner} variant that applies a single CNN feature extractor over the manually merged global map from the map merger, instead of forcing the network to learn to fuse each agent's information.
\end{itemize}

As shown in Fig.~\ref{fig:ab-scp}, the full {\name} module produces the lowest \emph{Steps} and \emph{Mutual Overlap}.
Among all the {\planner} variants, \emph{{\planner} w.o. Act. Gen.} produces the highest \emph{Steps}. This suggests that a simple Gaussian representation of actions may not be able to fully capture the distribution of good long-term goals, which can be highly multi-modal in the early exploration stage. In scene \emph{Dryville}, \emph{{\planner} w.o. TeamFormer} performs much worse and shows larger training unstability than the full model, showing the importance of jointly leveraging intra-agent and spatial relationships in a hierarchical manner. In addition, \emph{{\planner}-merge} produces a very high \emph{Mutual Overlap} in scene \emph{Dryville}. We hypothesis that this is due to the fact that many agent-specific information are lost in the manually merged maps while {\planner} can learn to utilize these features implicitly. 


\subsubsection{Ablation Study on Spatial-TeamFormer}
\label{sec: ST}
We consider the following variants of {\name} by altering the components of Spatial-TeamFormer as follows:



\begin{itemize}
    \item \textbf{No Ind. Spatial Enc.}: Individual Spatial Encoder is removed from Spatial-TeamFormer
    \item \textbf{No Team Rel. Enc.}: Similarly, this variant removes Team Relation Encoder while only keeps Individual Spatial Encoder.
    \item \textbf{Unified}: 
     This variant applies a single unified transformer over the spatial features from all the agents instead of the hierarchical design in Spatial-TeamFormer. In particular, we directly feed all the $N\times G\times G$ features into a big transformer model to generate an invariant representation. 
    
    
    \item \textbf{Flattened}: 
    In this variant, we do not keep the spatial structure of feature maps. Instead, we first convert the CNN extracted feature into a flatten vector for each agent and then simply feed these $N$ flattened vectors to a standard transformer model for feature learning. We remark that this variant is exactly the same as~\citep{multiagent-navigation1}.
    
\end{itemize}

\begin{figure*}[h]
	\centering

    \includegraphics[width=0.9\linewidth]{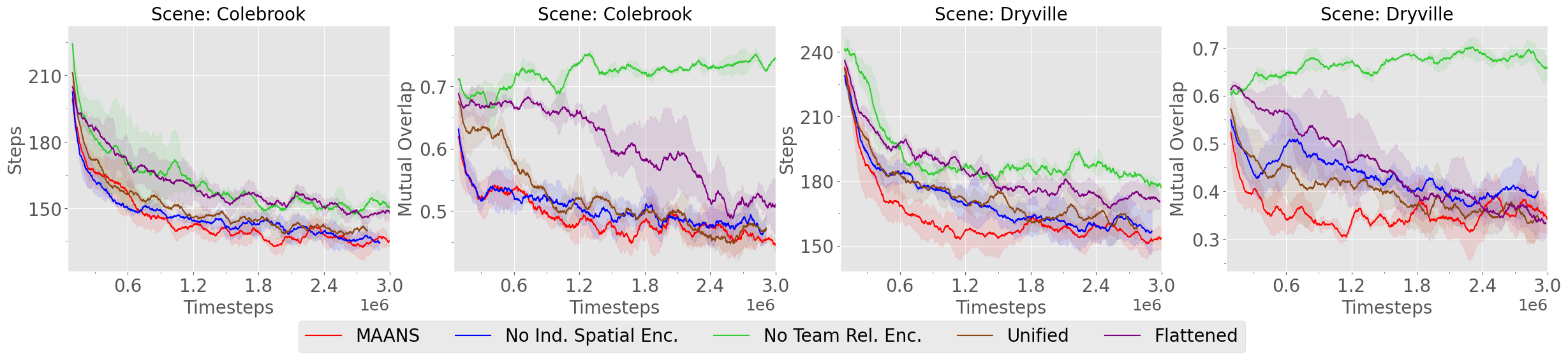}

	\centering 
	\caption{Ablation studies on Spatial-TeamFormer.}
\label{fig:baselines}

\end{figure*}

 We report training curves in \cref{fig:baselines}.
Compared with Spatial-TeamFormer, \emph{No Team Rel. Enc.} has the highest \emph{Mutual Overlap} and worst \emph{Steps} on both scenes, which suggests that lacking partners' relationship attention significantly lowers the cooperation efficiency. We remark that \emph{No Team Rel. Enc.} is indeed a single-agent variant of Spatial-TeamFormer: each agent plans global goal using its individual information while doing path planning still with the merged global map. The variant using \emph{Flattened} features is also performing much worse than the full model with a clear margin, showing that the network architecture without utilizing spatial inductive bias could hurt final performance. When individual spatial encoder is removed (\emph{No Ind. Spatial Enc.}), the sample efficiency drops greatly in scene \emph{Dryville} and the method achieves a higher \emph{Mutual Overlap} than {\name}. \emph{Unified} also has worse sample efficiency that the full model. Note that \emph{Unified} shows greater performance than \emph{Flattened}, again confirming the importance of utilizing spatial inductive bias.

\subsection{Main Results}

Due to space constraints, we only present a selected portion of the most competitive results in the main paper and defer the full results to Appendix \ref{app: add}. 


\subsubsection{Comparison with Planning-based Baselines and RL baseline}
\textbf{(1) Training with a Fixed Team Size:}
We first report the performance of {\name} and selected the baseline methods with a fixed team size of $N=2$ agents on both representative training scenes and testing scenes in~\cref{tab: training_agents_plan}. We remark that only 9 policies of representative training scenes are used to do policy distillation(PD) since it takes a lot of work to train a separated policy for each scene. Except for the \emph{Mutual Overlap} on the testing scenes where the performance is slightly worse than \emph{Voronoi}, {\name} still outperforms all planning-based baselines in \emph{Steps} and \emph{Coverage} metrics on training and testing scenes. More concretely, {\name} reduces 20.56\% exploration steps on training scenes and 7.99\% exploration steps on testing scenes than the best planning-based competitor. We also compare with an RL baseline \emph{{\name} w.o. PD}, which is trained by randomly sampling all training scenes instead of policy distillation. \emph{{\name} w.o. PD} performs much worse than {\name} on both training and testing scenes, and slightly worse than the best single-agent planning-based method \emph{RRT} and multi-agent planning-based method \emph{Voronoi} on testing scenes, indicating the necessity of introducing policy distillation. 

We also observe that APF and WMA-RRT, as multi-agent planning baselines, even perform worse than single-agent methods. We empirically found this is due to the formally-designed cooperation paradigm they adopt, which imposes great restriction to agents’ behaviors.
In contrast, {\name}, by using {\planner}, could perform more complicated cooperative strategy to fully explore the scene. Further illustration and analysis could be found in Appendix \ref{app: case}.

\begin{table*}
\centering
\tiny
\begin{tabular}{crcccccc|c} 
\toprule
Sce.                 & Metrics   & Utility      & RRT      & APF     &  WMA-RRT & Voronoi & {\name} w.o PD & {\name}          \\ 
\midrule                                  
\multirow{3}{*}{\tabincell{c}{Train}}

                         & \emph{Mut. Over.} $\downarrow$    & 0.68(0.01)   & 0.53(0.02)  & 0.61(0.01) &  0.61(0.01) &0.44(0.01) & 0.46(0.01) & \textbf{0.42(0.01)}  \\ 
\cmidrule{2-9}
                         & \emph{Steps} $\downarrow$  & 236.15(3.61) & 199.59(3.27)  & 251.41(3.15) & 268.20(2.24)& 237.04(2.95) &180.25(2.35) & \textbf{158.55(2.25)}  \\ 
\cmidrule{2-9}
                         & \emph{Coverage} $\uparrow$ & 0.92(0.01)   & 0.96(0.00)  & 0.90(0.01)  &0.87(0.01) &0.93(0.00)   &0.96(0.00) & \textbf{0.97(0.00)}  \\ 
\midrule
 \multirow{3}{*}{\tabincell{c}{Test}} 
                         & \emph{Mut. Over.}$\downarrow$&
                         0.69(0.01) &   0.57(0.01) &   0.57(0.01) &   0.64(0.01) &   \textbf{0.51(0.01)} &   0.57(0.01)   &   0.54(0.02)   \\ 
\cmidrule{2-9}
                         & \emph{Steps} $\downarrow$ & 161.28(2.32) & 157.29(2.59) & 181.18(4.17) & 198.92(3.83) & 156.68(3.21) &159.53(2.73)  &     \textbf{144.16(2.52)} \\ 
\cmidrule{2-9}
                         & \emph{Coverage} $\uparrow$ & 0.95(0.00) &   0.95(0.01) &   0.93(0.01) &   0.91(0.01) &   \textbf{0.96(0.01)} &  \textbf{0.96(0.00)}   &    \textbf{0.96(0.00)}  \\ 
\bottomrule
\end{tabular}
\vspace{1mm}
\caption{Performance of {\name} and selected \emph{planning-based} baselines and RL baseline with a fixed size of $N=2$ agents on both training and testing scenes.}
\label{tab: training_agents_plan}
\end{table*}

\textbf{(2) Zero-Shot Transfer to Different Team Sizes:}
In this part, we directly apply the policies trained with $N=2$ agents to the scenes of $N=3,4$ agents respectively. The zero-shot generalization performance of {\name} compared with the best single-agent baseline \emph{RRT} and the best multi-agent baseline \emph{Voronoi} on both training and testing scenes is shown in~\cref{tab: unseen_and_varying_agents_plan}. Note that experiments on testing scenes are extremely challenging since {\name} is never trained with $N=3,4$ team sizes on testing scenes.
Although {\name} is only trained on the team size of $2$ on training scenes, {\name} achieves much better performance than the best planning-based methods with every novel team size on training scenes (17.56\% fewer \emph{Steps} with $N=3$ and 18.36\% fewer \emph{Steps} with $N=4$) and comparable performance on testing scenes ($<3$ more \emph{Steps} with $N=3,4$). 




\begin{table*}
\centering
\vspace{-5mm}
\tiny
\begin{tabular}{ccccc|ccc} 
\toprule
\multirow{2}{*}{\begin{tabular}[c]{@{}c@{}}\\\# Agent\end{tabular}} & \multirow{2}{*}{Metrics}         & \multicolumn{3}{c|}{Training Scenes}                                                                 & \multicolumn{3}{c}{Testing Scenes}                                            \\ 
\cmidrule{3-8}
                                                                    &                                  & RRT          & Voronoi             &  {\name}                                           & RRT          & Voronoi                        &              {\name}                 \\ 
\midrule
\multirow{3}{*}{3}                                                  & \textit{Mut. Over.} $\downarrow$ & 0.44(0.01)   & \textbf{0.37(0.01)} & \multicolumn{1}{c|}{0.42(0.01)}                                 & 0.45(0.01)   & \textbf{0.43(0.01)}            & 0.53(0.01)                    \\ 
\cmidrule{2-8}
                                                                    & \textit{Steps} $\downarrow$      & 155.13(3.26) & 180.27(2.51)        & \multicolumn{1}{c|}{\textbf{127.88(1.91)}}                      & 128.33(1.66) & \textbf{\textbf{119.98(2.31)}} & 122.48(2.22)                  \\ 
\cmidrule{2-8}
                                                                    & \textit{Coverage} $\uparrow$     & 0.95(0.01)   & 0.95(0.00)          & \multicolumn{1}{c|}{\textbf{0.97(0.00)}}                        & 0.95(0.01)   & \textbf{\textbf{0.96(0.00)}}   & \textbf{\textbf{0.96(0.00)}}  \\ 
\midrule
\multirow{3}{*}{4}                                                  & \textit{Mut. Over.} $\downarrow$ & 0.36(0.01)   & \textbf{0.34(0.01)} & \multicolumn{1}{c|}{0.42(0.01)}                                 & 0.41(0.01)   & \textbf{0.39(0.01) }           & 0.50(0.01)                    \\ 
\cmidrule{2-8}
                                                                    & \textit{Steps} $\downarrow$      & 140.57(1.78) & 147.01(2.38)        & \multicolumn{1}{c|}{\textbf{114.75(1.69)}}                      & 111.30(1.58) & \textbf{101.90(2.36)}          & 109.07(2.02)                  \\ 
\cmidrule{2-8}
                                                                    & \textit{Coverage} $\uparrow$     & 0.92(0.01)   & 0.93(0.00)          & \textbf{0.96(0.00)}                                             & 0.93(0.01)   & \textbf{0.95(0.00) }           & 0.94(0.00)                    \\ 
\midrule
\multirow{3}{*}{2 $\Rightarrow$ 3}                                 & \textit{Mut. Over.}$\downarrow$  & 0.36(0.01)   & 0.35(0.01)          & \textbf{0.30(0.01)}                                             & 0.43(0.01)   & \textbf{0.32(0.02) }                    & 0.46(0.01)                    \\ 
\cmidrule{2-8}
                                                                    & \textit{Steps}$\downarrow$       & 185.94(1.83) & 200.91(2.32)        & \begin{tabular}[c]{@{}c@{}}\textbf{148.82(2.01)}\\\end{tabular} & 136.42(2.41) & 148.12(6.69)                   & \textbf{134.11(2.88) }                 \\ 
\cmidrule{2-8}
                                                                    & \textit{Coverage}$\uparrow$      & 0.94(0.00)   & 0.92(0.00)          & \textbf{0.96(0.00)}                                                      & \textbf{0.96(0.01)}   & 0.92(0.05)                     & \textbf{0.96(0.00)}                    \\ 
\midrule
\multirow{3}{*}{3 $\Rightarrow$ 2}                                 & \textit{Mut. Over.}$\downarrow$  & 0.35(0.01)   & \textbf{0.33(0.01)}          & 0.41(0.01)                                                      & \textbf{0.39(0.01)}   & 0.42(0.01)                     & 0.43(0.01)                    \\ 
\cmidrule{2-8}
                                                                    & \textit{Steps}$\downarrow$       & 187.93(1.98) & 206.94(2.50)        & \textbf{145.14(2.83)}                                                   & 139.52(3.74) & \textbf{133.77(2.83) }                  & 145.43(3.44)                  \\ 
\cmidrule{2-8}
                                                                    & \textit{Coverage}$\uparrow$      & 0.91(0.00)   & 0.89(0.01)          & \textbf{0.95(0.00)}                                                      & 0.94(0.01)   &\textbf{ 0.95(0.01)   }                  & 0.94(0.01)                    \\
\bottomrule
\end{tabular}
\vspace{2mm}
\caption{Generalization performance of {\name} and selected \emph{planning-based methods} to novel fixed and varying team sizes on training and testing scenes. Note that {\name} has the best performance on training scenes and comparable results on testing scenes.}
\label{tab: unseen_and_varying_agents_plan}
\end{table*}

\textbf{(3) Varying Team Size within an Episode}
We further consider the setting where the team size varies within an episode. 
We summarize the zero-shot generalization performance of {\name} compared with two selected planning-based methods RRT and Voronoi in~\cref{tab: unseen_and_varying_agents_plan}. We use "$N_1\Rightarrow N_2$" to denote that each episode starts with $N_1$ agents and the team size immediately switches to $N_2$ after $90$ timesteps. Note that {\name} is trained on the training scenes with fixed team size, the varying team size setting is a zero-shot generalization challenge for {\name}.
In cases where the team size increases, {\name} produces substantially better performances w.r.t. every metric. 
In particular, {\name} achieves 33 fewer \emph{Steps} in training scenes and lower \emph{Steps} than other methods in testing scenes, which suggests that {\name} has the capability to adaptively adjust its strategy.
Regarding the cases where the team size decreases, {\name} consumes over $40$ fewer $Steps$ in $3\Rightarrow 2$ than RRT in training scenes. 

We remark that decreasing the team size is particularly challenging since the absence of some agents might immediately leave a large part of the house unexplored and consequently, the team should immediately update their original plan with drastically different goal assignments. 



\subsection{Learned Strategy}

\begin{figure}[h]
	\centering
    \includegraphics[width=0.6\textwidth]{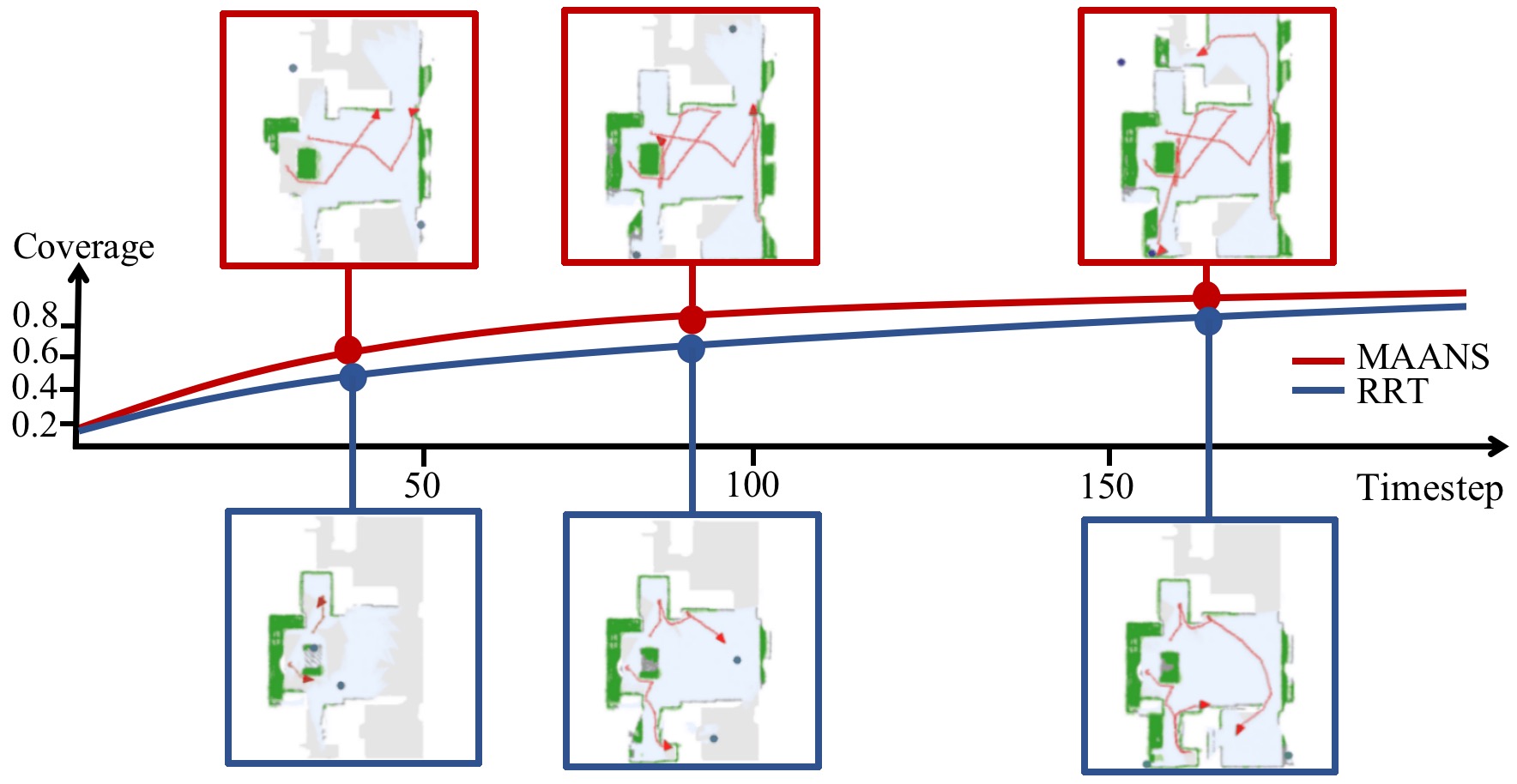}
	\centering 
	\caption{Learned strategy on scene \emph{Colebrook} of {\name} vs. RRT, where the red line with arrow represents the trajectory, the explored area shows in blue and the obstacle shows in green. {\name} achieves much higher and faster coverage ratio than RRT throughout the episode.}
\label{fig:case-study}

\end{figure}

\cref{fig:case-study} demonstrates two 2-agent trials of {\name} and RRT, the most competitive planning-based method, with the same birth place. The merged global map are shown in keep timesteps. As shown in Fig~\ref{fig:case-study}, {\name}'s coverage ratio goes up faster than RRT, indicating higher exploration efficiency. At timestep around $90$, {\name} produces global goals successfully allocate the agents towards two distant unexplored area while RRT guides the agents towards the same part of the map. And at timestep around $170$ when {\name} reaches $90\%$ coverage ratio, RRT still stuck in previous explored area though there is obviously another large open space. Notice that at this key timestep RRT selects two frontiers that are marked unexplored but with no actual benefit, which an agent utilizing prior knowledge about room structures would certainly avoid.

\section{Conclusion}
\label{sec:conclusion}

We propose the first multi-agent cooperative exploration framework, \emph{Multi-Agent Active Neural SLAM} (\name) that outperforms planning-based competitors in a photo-realist physical environment. The key component of {\name} is the RL-based planning module, \emph{Multi-agent Spatial Planner (\planner)}, which leverages a transformer-based architecture, Spatial-TeamFormer, to capture team-size-invariant representation with strong spatial structures. We also implement a collection of multi-agent-specific enhancements and 
policy distillation for better generalization. Experiments on Habitat show that {\name} achieves better training and testing performances than all the baselines. We hope {\name} can inspire more powerful multi-agent methods in the future.

\newpage
\appendix

We would suggest to visit \url{https://sites.google.com/view/maans} for more information.

\section{{\name} Details}
Multi-Agent Active Neural SLAM (\name) consists of 4 modules (1) Neural SLAM; (2) Map Refiner and Map Merger; (3) Local Policy and Local Planner; (4) Multi-agent Spatial Planner ({\planner}). Here we describe each module in detail.

\subsection{Neural SLAM}
The Neural SLAM Module for map reconstruction and pose estimation and the Local Policy for action output in our work are directly derived from ANS~\citep{ans}. Neural SLAM Module trained by supervised learning provides each agent an updated reconstructed map individually at every timestep. In order to recover a metric map with high accuracy, Neural SLAM Module takes as input current RGB observation $o_t$, current and last pose $x'_{t-1:t}$ from sensors, last pose estimation $\hat{x}_{t-1}$ and last map prediction $\hat{m}_{t-1}$, and outputs a map prediction $\hat{m}_t$ and a pose estimation $\hat{x}_t$, where $t$ represents the current timestep. Note that noises are introduced in simulation to mimic realistic situations. 

\subsection{Map Refiner and Map Merger}
For a better choice of cooperative global goals, we designed a Map Refiner for arranging all maps into the same coordinate system and a Map Merger for shared map reconstruction. More concretely, Map Refiner obtains the egocentric \emph{global} map from a series of past egocentric \emph{local} maps and unifies the global maps from all agents in the same coordinate system based on the pose estimates. Besides, there is a dilemma that the egocentric local map contains part of redundant space if an agent reaches the edge of the house, resulting in a large portion of invisible region around the explorable area. To promise an effective CNN feature extraction and more accurate global goal generation, we clip the unexplorable boundary and enlarge the explorable region.

Map Merger leverages all enlarged global maps from the Map Refiner to compose a shared map through max-pooling operator for each pixel location, which indicates the probability of being explored or the obstacle. As a result, the local planner produces the sub-goals on the merged global map, which is much more informative. Note that the merged map is merely employed in local planner to plan path, but not introduced in {\planner}, which only utilizes agents' egocentric global maps to infer global goals.

\subsection{Local Planner and Local Policy}

To effectively reach a global goal, the agent first plans a path to this long-term goal in a manually merged global map using Local Planner, which is mainly based on Fast Marching Method (FMM)~\citep{fmm}, and generates a sequence of short-term sub-goals. 
The Local Policy learns to produce next action via imitation learning. The input of the Local Policy includes the relative angle and distance from the current position to the short-term goal as well as current RGB observations.

\subsection{Multi-agent Spatial Planner}
\label{app: msp}
\subsubsection{Input Representation}

The shared CNN Feature Extractor in Multi-agent Spatial Planner firstly takes in a $240\times 240$ map with 6 channels as input, containing

\begin{itemize}
    \item Obstacle channel: indicating the likelihood of being an obstacle of each pixel
    \item Explored region channel: denoting the probability of being explored of each pixel
    \item One-hot position channel: describing the position of the agent with an one-hot metric map.
    \item Trajectory channel: expressing the history trace of each agent with exponentially decaying weight to emphasize the direction of the trace: 
    \begin{equation*}
        \begin{array}{l}V_{x,y}^t=\left\{\begin{array}{l}
        \begin{array}{l}1\quad \quad\quad\text{current position}\\\varepsilon V^{t-1}_{x,y}\quad \text{otherwise}\end{array}\end{array} \right.
        \\
        \end{array}
    \end{equation*}
    where $V^t$ denotes the trajectory channel at timestep $t$.
    \item One-hot global goal channel: demonstrating the position of the last global goal in an one-hot manner.
    \item One-hot Goal history channel: recording all the previous global goals of the agent.
\end{itemize}

Besides CNN spatial maps, we also introduce additional features, including agent-specific embeddings of its identity and current position and grid features, i.e., the embeddings of the relative coordinate of each grid to the agent position as well as the embedding of the previous global goal. 

\begin{itemize}
    \item Position Embedding: described as trainable ${G\times G\times D}$ parameters as part of the neural network. Two types of position embeddings are used to distinguish from the decision-making agent and it's partners. Note that $G$ and $D$ is respectively 8 and 128.
    \item Relative Coordinate Embedding: describing the relative position of the agent with the $G\times G$ coarse-grained maps.
    \item Previous Global Goal Embedding: expressing the relative position of last global goal with the $G\times G$ coarse-grained maps.
\end{itemize}

These embeddings are all concatenated with features outputted by CNN feature extractors and then fed into {\planner}.

\subsubsection{Hierarchical Action Space} 

{\name} adopts a hierarchical action space to represent global goals, where a global goal is consequently composed of a high-level discrete \emph{region} $g=(g_x,g_y)$ and a low-level fine-grained continuous \emph{point} $p=(p_x,p_y)$. To be more specific, the whole world-frame occupancy map is discretized into $8\times 8$  uniform regions and a global goal $(x_l,y_l)$ is decomposed into two levels,
\begin{align*}
    x_l&=\frac{g_x+p_x}{8}\\
    y_l&=\frac{g_y+p_y}{8}
\end{align*}
In {\planner}, the region head outputs a 64-dim vector denoting the categorical distribution of $g$, while the point head outputs a bivariate Gaussian distribution $\mathcal N(\mu,\Sigma)$. Point $p$ is obtained by applying Sigmoid to the Gaussian random variables, i.e.,
\begin{align*}
    \hat p&=(\hat p_x,\hat p_y)\sim \mathcal N(\mu,\Sigma)\\
    p_x&=\text{Sigmoid}(\hat p_x)\\
    p_y&=\text{Sigmoid}(\hat p_y)
\end{align*}

\subsubsection{Reward Function}
We design the reward in a team-based fashion, comprising of the coverage reward, the success reward, the overlap penalty and the time penalty. 
For a unified representation, $Ratio^t$ indicates the total coverage ratio at timestep $t$, $Area^t$ is the the total coverage area at timestep $t$, and $Area^t_k$ represents the explored area of agent $k$. The details of 4 kinds of reward gained by agent $k$ at timestep $t$ are listed as below.
\begin{itemize}
    \item \textbf{Coverage Reward:} The coverage reward is a combination of the team coverage reward and the individual coverage reward. Team coverage reward illustrates the increment of explored area at timestep $t$, and is proportional to $\Delta Area^t=Area^t-Area^{t-1}$. For the consideration of individual contribution to the whole team exploration, the individual coverage reward is proportional to $\Delta Area^t_k=Area^t_k-Area^{t-1}_k\bigcap Area^{t-1}$. We remark that $\Delta Area^t_k \neq Area^t_k-Area^{t-1}_k$, which suppresses cooperation but leads to the individual exploration. The coefficient is $0.02$.
    \item \textbf{Success Reward:} To encourage agent $k$ achieves the target coverage ratio as much as possible, we give the bonus $1\cdot Ratio^t$ to the agent if the total coverage rate is reached in $95\%$, and $0.5\cdot Ratio^t$ when $90\%$ coverage rate is realized.
    \item \textbf{Overlap Penalty:} The overlap penalty is applied to reduce repetitive exploration among agents so as to enhance cooperation capability. It is described as:
    \begin{equation*}
        \begin{array}{l}\left\{\begin{array}{l}
        \begin{array}{l}-\Delta A_{overlap}^t\times0.01,\;Ratio^t\;<0.9\\-\Delta A_{overlap}^t\times0.006,0.9\leq\;Ratio^t<0.95\\0,\;Ratio^t\ge0.95\end{array}\end{array} \right.
        \\
        \end{array}
    \end{equation*}
    $\Delta A_{overlap}^t=A_{overlap}^t-A_{overlap}^{t-1}$ denotes the increment of the average overlapped explored area $A_{overlap}^t$ between each two agents at timestep $t$. In practical, the $A_{overlap_{k,u}}^t=Area_k^{t}\bigcap Area_u^{t-1}$ between agent $k$ and agent $u$ transforms into the sum of a one-hot map, where the grid on the map will be valued when its total value of two agents' explored probability is greater than $1.2$.    
    \item \textbf{Time Penalty:} For the sake of an efficient exploration, we propose the time penalty:
    \begin{equation*}
        \begin{array}{l}\left\{\begin{array}{l}
        \begin{array}{l}-0.002,\;Ratio^t\;<0.9\\-0.001,0.9\leq\;Ratio^t<0.95\\-0.0002,\;0.95\leq Ratio^t<0.97\end{array}\end{array} \right.
        \\
        \end{array}
    \end{equation*}
\end{itemize}

The linear combination of four parts is the final team-based reward. Note that all the explored and obstacle maps are in the scale of $5cm$ for each grid and the measurement unit of all the area is $m^2$.

\subsubsection{Architecture}

CNN feature extractor is composed of 5 convolution layers as well as max pooling layers.  Hyperparameters of these layers are listed in \cref{tab:cnn}. Except the last layer, each layer is followed by a max pooling layer with kernel size $2$.

The core part of {\planner}, Spatial-Teamformer, contains several blocks, each of which consists of two layers, i.e. an Individual Spatial Encoder and a Team Relation Encoder. Both Individual Spatial Encoder and Team Relation Encoder are self-attention layers with residual connection. We remark that the self-attention mechanism in {\planner} is exactly the same as transformer encoder in \citep{vit}. Hyperparameters of Spatial-Teamformer are listed in \cref{tab:ST}. 

\begin{table}[bt]
\centering
\begin{tabular}{ccccc}
\toprule
Layer &                      Out Channels &         Kernel Size & Stride & Padding \\
\midrule
1 & 32 & 3 & 1 & 1 \\
    2 &  64 & 3 & 1 & 1 \\
3 & 128 & 3 & 1 & 1 \\
    4 & 64 & 3 & 1 & 1 \\
  5 & 32 & 3 & 2 & 1 \\
\bottomrule
\end{tabular}
\caption{CNN feature extractor hyperparameters}
\label{tab:cnn}
\end{table}

\begin{table}[bt]
\centering
\vspace{1mm}
\begin{tabular}{cc}
\toprule
hyperparameters      & value  \\
\midrule
    \# of attention heads & 4 \\
    attention head dimension & 32\\
    attention hidden Size & 128\\
    \# of Spatial-Teamformer blocks & 2\\
\bottomrule
\end{tabular}
\caption{{\planner} hyperparameters}
\label{tab:ST}
\vspace{-5mm}
\end{table}




\section{Baselines}
\label{app: baseline}

We implemented 6 classical planning-based methods, including 3 single-agent methods and 3 multi-agent baselines.

\subsection{Single-agent Baselines}

\begin{itemize}
\item \textbf{Nearest}~\citep{frontier1} selects the nearest frontier as global goal~\citep{frontier2} via breadth first search on the merged global map.
\item \textbf{Utility}~\citep{Utility} chooses the frontier with the largest information gain~\citep{frontier3}.
\item \textbf{RRT}~\citep{RRT} generates a collision-free random tree rooted at the agent's current location. After collecting enough tree nodes that lie on unexplored region, i.e. frontiers, RRT chooses the one with the highest utility $u(p)=IG(p)-N(p)$, where $IG(p)$ and $N(p)$ are respectively the normalized information gain and navigation cost of $p$. Pseudocode of RRT is shown in \cref{algo:RRT}. In each iteration, a random point $p$ is draw and a new node $t$ is generated by expanding from $s$ to $p$ with distance $L$, where $s$ is the closest tree node to $p$. If segment $(s,t)$ has no collision with obstacles in $M$, $t$ is inserted into the target list or the tree according to whether $t$ is in unexplored area or not. Finally, the goal is chosen from the target list with the largest utility $u(c) =IG(c)-N(c)$ where $IG(c)$ is the information gain and $N(c)$ is the navigation cost. $IG(c)$ is computed by the number of unexplored grids within $1.5m$ to $c$, as mentioned above. $N(c)$ is computed as the euclidean distance between the agent location and point $c$. To keep these two values at the same scale, we normalize $IG(\cdot)$ and $N(\cdot)$ to $[0,1]$ w.r.t all cluster centers. 

\end{itemize}

\subsection{Multi-agent Baselines}

\begin{itemize}
\item \textbf{APF}~\citep{APF} first computes a potential field $F$ based on explored occupancy map and current agent locations, and then follows the fastest descending direction of $F$ to find a frontier as the global goal. Resistance force among agents is introduced in APF to avoid repetitive exploration. Pseudocode of APF is provided in \cref{algo:APF}. Line 6-12 computes the resistance force between every pair of agents where $D$ is the influence radius. In line 13-18, distance maps starting from cluster centers are computed and the corresponding reciprocals are added into the potential field so as one agent approaches the frontier, the potential drops. Here $w_c$ is the weight of cluster $c$, which is the number of targets in this cluster. Consequently an agent would prefer to seek for frontiers that are closer and with more neighboring frontiers. Line 20-25 shows the process to find the fastest potential descending path, at each iteration the agent moves to the cell with the smallest potential among all neighboring ones. $T$ is the maximum number of iterations and $C_{repeat}$ is repeat penalty to avoid agents wandering around cells with same potentials.

\item \textbf{WMA-RRT}~\citep{WMA-RRT} WMA-RRT is a multi-agent variant of RRT. Pseudocode of WMA-RRT is provided in \cref{algo:wma-rrt}. By maintaining a rooted tree together, agents share information to finish exploration. To impose cooperation using the shared tree, WMA-RRT uses a locking mechanism to avoid agents exploring same part of the tree and restricts agents to walking along the edge of the tree to ensure a strict system. Agents choose a node in the tree as a global goal and mark whether a subtree has been completedly searched. Although this is a multi-agent variant of RRT, we empirically found it perform much worse than RRT. We found this is because the locking mechanism actually restricts agent behaviors greatly and the algorithm itself is incompatible with active SLAM. For the former, agents are often forced not to explore large open areas which require multi-agent effort because locked by another agent. For the latter, WMA-RRT was originally designed for the case a ground-truth mapping is given and adding new nodes into the tree during exploration would cause mis-labeled completed subtree. Also, WMA-RRT do not perform value estimation over the nodes, making agents committed to branches that do not increase coverage much. Therefore, though WMA-RRT is guaranteed to reach fully coverage, it's inherently not suitable for the setting to maximize coverage ratio. As a comparison between RRT and WMA-RRT, WMA-RRT only utilizes RRT to expand tree while perform multi-agent planning using a strict tree-search procedure while RRT uses the random tree equipped with utility estimation to do planning. For detailed description of the algorithm, we encourage readers to check out ~\citep{WMA-RRT}.

\item \textbf{Voronoi}~\citep{Voronoi} The voronoi-based method first partitions the map via voronoi partition and assigns components to agents so that each agent owns parts that are closest to it. Then each agent finds its own global goal by finding a frontier point with largest potential as in Utility within its own partition. In this way, duplicated exploration is be avoided. Pseudocode of Voronoi is provided in \cref{algo:voronoi}.
\end{itemize}

Finally, we also evaluate the performance of random policies for references. 
To eliminate negative impact of visual blind area, the area within a distance of $2.5m$ to the agent is virtually marked as explored when choosing frontiers so that these baselines would output far enough global goals. The number of unexplored grids within a distance of $1.5m$ to a frontier $p$ is defined as the information gain. All these baselines regenerate new global goals every $15$ time steps, which is consistent with {\planner}. Case studies and failure modes of these methods are provided in Section \ref{app: case}.

\begin{algorithm}
    \caption{Rapid-exploring Random Tree (RRT)}
    \label{algo:RRT}
    \begin{algorithmic}[1]
        \REQUIRE Map $M$ and agent location $loc$.
        \ENSURE Selected frontier goal
        \STATE $NodeList\leftarrow \{loc\}, Targets\leftarrow\{\}$
        \STATE $i\leftarrow 0$
        \WHILE{$i<T$ and $|Targets|<N_{target}$}
            \STATE $i\leftarrow i+1$
            \STATE $p\leftarrow $ a random point
            \STATE $s\leftarrow \arg\min_{u\in NodeList}||u-p||_2$
            \STATE $t\leftarrow Steer(s, p, L)$
            \IF{$No\_Collision(M, s, t)$}
                \IF{$t$ lies in unexplored area}
                    \STATE $Targets\leftarrow Targets + \{t\}$
                \ELSE
                    \STATE $NodeList\leftarrow NodeList + \{t\}$
                \ENDIF
            \ENDIF
        \ENDWHILE
        \STATE $C\leftarrow $ clusters of points in $Targets$.
        \STATE $goal\leftarrow \arg\min_{c\in C} IG(c) - N(c)$
        \RETURN $goal$
    \end{algorithmic}
\end{algorithm}

\begin{algorithm}
    \caption{Artificial Potential Field (APF)}
    \label{algo:APF}
    \begin{algorithmic}[1]
        \REQUIRE Map $M$, number of agents $n$ and agent locations $loc_{1}\dots loc_{n}$.
        \ENSURE Selected goals for each agent
        \STATE $P\leftarrow$ frontiers in $M$
        \STATE $C\leftarrow$ clusters of frontiers $P$
        \STATE $goals\leftarrow$ an empty list
        \FOR{$i=1\rightarrow n$}
            \STATE $F\leftarrow$ zero potential field
            \STATE // Compute Resistance force
            \FOR{$j=1\rightarrow n$}
                \FOR{unoccupied grid $p\in M$}
                    \IF{$j\neq i$ and $||p-loc_j||_2<D$}
                        \STATE $F_p \leftarrow F_p + k_D\cdot(D-||p-loc_j||_2)$
                    \ENDIF
                \ENDFOR
            \ENDFOR
            \FOR{$c\in C$}
                \STATE Run breadth-first search to compute distance map $dis$ starting from $c$
                \STATE $F\leftarrow F - dis^{-1}\cdot w_c$
            \ENDFOR
            \STATE $u\leftarrow loc_i, cnt\leftarrow 0$
            \WHILE{$u\notin M$ and $F_u$ is not a local minima and $cnt<T$}
                \STATE $cnt\leftarrow cnt+1$
                \STATE $F_u\leftarrow F_u+C_{repeat}$
                \STATE $u\leftarrow\arg\min_{v\in Neigh(u)}F_v$
            \ENDWHILE
            \STATE{append $u$ to the end of $goals$}
        \ENDFOR
        \RETURN $goals$
    \end{algorithmic}
\end{algorithm}

\begin{algorithm}
    \caption{Weighted Multi-Agent RRT}
    \label{algo:wma-rrt}
    \begin{algorithmic}[0]
        \STATE \textbf{Find\_Next\_Point($a$)}:
        \IF {Node $a$ is a leaf node}
            \RETURN $a$        
        \ENDIF
        \FOR{edge $e = (a, b)\in Child(a)$ in clock-wise order}
            \IF {$e$ is not locked and subtree rooted at node $b$ is not completed}
                \STATE Lock edge $e$ for $LockTime$ timesteps
                \RETURN Find\_Next\_Point($b$)
            \ENDIF
        \ENDFOR
        \STATE Mark subtree rooted at $a$ as completed
        \RETURN Parent node of $a$
        \STATE ~
        \STATE \textbf{Main():}
        \STATE Reset the environment $env$
        \STATE PHASE $\leftarrow$ "Gather Stage"
        \STATE Initialize rooted tree $T=\emptyset$
        \WHILE{$env$ is not done}
            \IF{agents are close enough}
                \STATE $RootLoc\leftarrow$  mean coordination of all agents
                \STATE Initialize root node $Root=Node(RootLoc)$
                \FOR{all agents $a=1\rightarrow n$ }
                    \STATE $T = T \cup \{(Root, Node(AgentLocation[a]))\}$        
                \ENDFOR
                \STATE PHASE $\leftarrow$ "Exploration Stage"
            \ENDIF
            \FOR{all agents $a=1\rightarrow n$ }
                \IF{PHASE = "Gather Stage"}
                    \STATE $goal_a\leftarrow$ mean coordination of all agents
                \ELSE
                    \STATE $goal_a\leftarrow $ \textbf{Find\_Next\_Goal($AgentNode[a]$)}
                \ENDIF
            \ENDFOR
            \STATE Compute directional action via $goal_1,\cdots,goal_N$
            \STATE Move one step and receiver observations
            \STATE Update global merged map
            \IF{PHASE = "Exploration Stage"}
                \STATE Add new nodes into $T$ given new global merged map using standard RRT
            \ENDIF
        \ENDWHILE
    \end{algorithmic}
\end{algorithm}

\begin{algorithm}
    \caption{Voronoi-based method}
    \label{algo:voronoi}
    \begin{algorithmic}[1]
        \REQUIRE Map $M$ and all agents' locations $loc_1,\cdots,loc_n$.
        \ENSURE Selected frontier goals for all agents
        \STATE Partition the map $M$ via voronoi parition into triangle blocks $B=\{b_1,\cdots,b_m\}$
        \STATE Initialize $P=[[],\cdots,[]]$, i.e. $n$ empty lists
        \FOR{all blocks $i=1\rightarrow m$}
            \STATE $dis_1,dis_2,\cdots,dis_n\leftarrow$ distance of all agents to block $b_i$.
            \STATE $a\leftarrow \arg\min_{a}dis_a$
            \STATE $P_a\leftarrow P_a\cup\{b_i\}$
        \ENDFOR
        \STATE $goal\leftarrow \emptyset$
        \FOR{all agents $a=1\rightarrow n$ }
            \STATE $goal_a\leftarrow$ frontier point within $P_a$ that has largest information gain.
        \ENDFOR
        \RETURN $goal$
    \end{algorithmic}
\end{algorithm}



\section{Evaluation Metrics}




We select 3 behavior statistics metric to show different characteristics of particular exploration methods.
\begin{itemize}
    \item \textbf{Coverage}: \emph{Coverage} represents the ratio of areas explored by the agents to the entire explorable space in the indoor scene at the end of the episode. Higher \emph{Coverage} implies more effective exploration. A cell of $5cm \times 5cm$ is considered explored/covered when the 2D projection on the floor of some depth image in the exploration history covers this cell.
    \item \textbf{Steps}: \emph{Steps} is the number of timesteps used by agents to achieve a coverage ratio of $90\%$ within an episode. Fewer \emph{Steps} implies faster exploration.
    \item \textbf{Mutual Overlap}: For effective collaboration, each agent should visit regions different from those explored by its teammates. We measure the average overlapping explored area over each pair of agents when the coverage ratio reaches 90\%, which we call \emph{Mutual Overlap}. \emph{Mutual Overlap} denotes the normalized value of mutual overlap. Lower \emph{Mutual Overlap} suggests better multi-agent coordination. 
\end{itemize}


\section{Training Details}

We adopt Multi-Agent Proximal Policy Optimization (MAPPO)~\citep{mappo}, a multi-agent extension of PPO, to train {\planner}. The pseudocode of MAPPO is provided in Algorithm~\ref{algo:mappo}. Detailed hyper-parameters are listed in \cref{tab:hyperparameter}.

\begin{algorithm}
    \caption{MAPPO}
    \label{algo:mappo}
\begin{algorithmic}
    \STATE  Initialize $\theta$, the parameters for policy $\pi$ and $\phi$, the parameters for critic $V$, using Orthogonal initialization (Hu et al., 2020) \; 
    \STATE Set learning rate $\alpha$
    \WHILE{$step \leq step_{\text{max}}$}
    \STATE set data buffer $D = \{\}$
    \FOR{$i = 1$ {\bfseries to} $num\_rollouts$}
    \STATE $\tau = []$ empty list
    \FOR{$t = 1$ {\bfseries to} $T$}
    \FORALL{agents $a$}
    \STATE $p_t^{(a)} = \pi(o_t^{(a)}; \theta)$
    \STATE $u_t^{(a)} \sim p_t^{(a)}$
    \STATE $v_t^{(a)} = V(s_t^{(a)}; \phi) $
    \ENDFOR 
    \STATE Execute actions $\boldsymbol{u_t}$, observe $r_t, s_{t+1}, \boldsymbol{o_{t+1}}$
    \STATE $\tau \pluseq [s_t, \boldsymbol{o_t}, \boldsymbol{u_t}, r_t, s_{t+1}, \boldsymbol{o_{t+1}}]$
    \ENDFOR
    \STATE Compute advantage estimate $\hat{A}$ via GAE on $\tau$
    \STATE Compute reward-to-go $\hat{R}$ on $\tau$ and normalize
    \STATE $D = D\cup \tau$
    \ENDFOR 
    \FOR{epoch $k=1,\dots,K$}
    \STATE $b \leftarrow$ sequence of random mini-batches from D with all agent data
    \FOR{batch $c$ in $b$}
    \STATE Adam update $\theta$ on $L(\theta)$ with batch $c$
    \STATE Adam update $\phi$ on $L(\phi)$ with batch $c$
    \ENDFOR
    \ENDFOR
    \ENDWHILE
\end{algorithmic}
\end{algorithm}
\vspace{-2mm}

\begin{table}[bt]
\centering
\begin{tabular}{cc}
\toprule
common hyperparameters      & value  \\
\midrule
gradient clip norm          & 10.0 \\
GAE lambda                   & 0.95    \\              
gamma                      & 0.99 \\
value loss                        & huber loss \\
huber delta                 & 10.0   \\
mini batch size           & batch size {/} mini-batch  \\
optimizer                & Adam       \\
optimizer epsilon            & 1e-5       \\
weight decay             & 0          \\
network initialization  & Orthogonal \\
use reward normalization   &True \\
use feature normalization   &True \\
learning rate & 2.5e-5 \\
parallel environment threads & 10 \\
number of local steps & 15\\
\bottomrule
\end{tabular}
\caption{MAPPO Hyperparameters}
\label{tab:hyperparameter}
\vspace{-5mm}
\end{table}

As for policy distillation, an expert network $\pi(g,x,y|s, \theta_{t}^{(i)})$ is trained, where $g$ is the region head, $(x,y)$ is the point head, $s$ is the state, for each training scenes and team size of $2$. After that, a student network is trained $\pi(g,x,y|s, \hat\theta)$ via performing behavior cloning from these expert networks. The student network is trained in dagger style: for each episode, we first collect data using $\pi(g,x,y|s,\hat\theta)$, and then run several updates to optimize the output of student using past experience. The objective function to minimize is sum of $L_g(\hat\theta)$, which is the KL divergence between the student region distribution $\pi_g(s,\hat \theta)$ and teacher region distribution $\pi_g(s,\theta_t^{(i)})$, and $L_{x,y}(\hat\theta)$, which is the square error loss between point head of student and that of teacher. Policy distillation uses Adam optimizer with $2.5e-5$ learning rate.

\vspace{-5mm}

\section{Experimental Setting}
\label{app: exp}
\subsection{Datasets}
We follow the dataset used in \emph{Active Neural SLAM} (ANS) \citep{ans}. The original Gibson Challenge dataset~\citep{xiazamirhe2018gibsonenv}, which could be used with Habitat Simulator, provides $72$ training and $14$ validation scenes. Note that Gibson testing set is not public but rather held on an online evaluation server for the PointGoal task, so the validation set is used as the testing instead of hyper-parameter tuning. We have made substantial efforts to check every single scene from the Gibson Challenge dataset and have to exclude a large portion of scenes not suitable for multi-agent exploration, including (i) scenes that have large disconnected regions; and (ii) scenes that have multiple floors (agents can not go upstairs) so that the agents are not possible to reach 90\% coverage of the entire house. Note that disconnected region is not an issue for semantic or point navigation tasks. 

We categorize the remaining scenes into $23$ training scenes, including 9 small scenes, 9 middle scenes and 5 large scenes based on explorable area, as well as $10$ testing scenes, which including 5 small scenes, 4 middle scenes and 1 large scene. Note that the validation set of the original Gibson Challenge dataset, i.e. the testing set, has 14 scenes, of which 8 scenes are eligible for our task, including 5 small scenes, 1 middle scene and 1 large scene. Since most scenes in the testing set are too small for multi-agent exploration task, we additionally add 3 middle scenes to the testing set. 
The common training paradigm for visual exploration is to randomly sample training scenes or team sizes at each training episode~\citep{singleagent-RL1}. However, we empirically observe that different Habitat scenes and team sizes may lead to drastically different exploration difficulties. 
During training, gradients from different configurations may negatively impact each other. Hence, our solution is to train a separate policy on each map and use policy distillation to extract a meta policy to tackle this problem.

\subsection{Assumption of Birth Position} We assume the agents has access to the \textbf{birth location} of each other while the locations during an episode are estimated using Neural SLAM module. The merged global map is fused using the estimated locations and the birth place.

\subsection{Episode Length}
First, we empirically found that as the number of agents grows, even random exploration can be particularly competitive (as shown in \cref{tab: unseen_agents_single}), which, we believe, is due to the limited explorable space of Gibson scenes.
Hence, we only test up to 4 agents and argue that the $2$-agent case is the most challenging.
Regarding the \emph{episode length}, it is estimated according to the number of timesteps when the strongest single-agent planning-based method, RRT, achieves 95\% coverage. 
Note that if the horizon is too long (e.g., 1000, which is used in ANS), almost all the methods will have the same the final coverage rate. 
In addition, the \emph{Mutual Overlap} and \emph{Steps} metrics are all estimated before the agents reach a 90\% coverage, which do not depend on the episode length. 
The choice of a higher re-planning frequency (e.g., re-plan per 15 timestep) is also due to a shorter episode horizon\footnote{All the agents will re-generate their personal long-term global goals using {\planner} in a synchronous manner every $15$ timesteps while ANS~\citep{ans} re-plans every 25 timesteps.}. 
All the baselines use the same planning frequency. 


\vspace{-2mm}
\section{Additional Experiment Results}
\label{app: add}
\subsection{Fixed Team Size}


\vspace{-1mm}
\subsubsection{Trained with Team size = 2}
We first report the performance of {\name} and all baseline methods with a fixed team size of $N=2$ agents on both 9 representative training scenes and unseen testing scenes in~\cref{tab: training_agents_single} and \cref{tab: training_agents_multi}.
{\name} outperforms all the planning-based baselines with a clear margin in every evaluation metric, particularly the \emph{Steps}, on both training and testing scenes. We can observe that \emph{Steps} of testing scenes is overall fewer than training scenes since we use 9 middle scenes for training while the testing set has more small scenes. And as the scene size grows, the performance of planning-based methods degrades a lot. We directly apply the policies trained with $N=2$ agents to the scenes of $N=3,4$ agents respectively. The zero-shot generalization performance of {\name} compared with all the baselines on 9 representative training scenes and testing scenes is shown in~\cref{tab: unseen_agents_single} and \cref{tab: unseen_agents_multi}. We can observe that {\name} still outperforms planning-based methods on training scenes. And {\name} could even achieve comparable performance on testing scenes despite {\name} having neither seen the testing scenes nor the novel team sizes.
\begin{table*}
\vspace{-5mm}
\centering
\scriptsize
\begin{tabular}{crccccc} 
\toprule
Sce.         & Metrics &   Random  & Nearest          & Utility      & RRT   & {\name}          \\ 
\midrule
\multirow{3}{*}{\tabincell{c}{Training\\Sce.}} 
                                                   & \emph{Mut. Over.} $\downarrow$ & 0.66(0.01)   & 0.53\scriptsize{(0.02)}   & 0.68\scriptsize{(0.01)}   & 0.53\scriptsize{(0.02)}          & \textbf{0.42\scriptsize{(0.01)}}    \\ 
\cmidrule{2-7}
                                                   & \emph{Steps} $\downarrow$      & 273.56(1.38) & 246.79\scriptsize{(3.90)} & 236.15\scriptsize{(3.61)} & 199.59\scriptsize{(3.27)}        & \textbf{158.55\scriptsize{(2.25)}}  \\ 
\cmidrule{2-7}
                                                   & \emph{Coverage} $\uparrow$     & 0.86(0.00)   & 0.91\scriptsize{(0.01)}   & 0.92\scriptsize{(0.01)}   & 0.96\scriptsize{(0.00)}          & \textbf{0.97\scriptsize{(0.00)}}    \\ 
\midrule
\multirow{3}{*}{\tabincell{c}{Testing\\Sce.}} 
                                                   & \emph{Mut. Over.} $\downarrow$ & 0.66(0.02) &   0.58(0.01) &   0.69(0.01) &   0.57(0.02) &         \textbf{0.54(0.02)   }         \\ 
\cmidrule{2-7}
                                                   & \emph{Steps} $\downarrow$      & 193.83(2.80) & 166.23(3.96) & 161.28(2.32) & 157.29(2.59) &       \textbf{144.16(2.52) } \\ 
\cmidrule{2-7}
                                                   & \emph{Coverage} $\uparrow$     & 0.93(0.00) &   0.95(0.00) &   0.95(0.00) &   0.95(0.01) &         \textbf{0.96(0.00)}    \\
\bottomrule
\end{tabular}
\caption{Performance of {\name} and \emph{single-agent baselines} with a fixed size of $N=2$ agents on both training and testing scenes.}
\label{tab: training_agents_single}
\vspace{-5mm}
\end{table*}
\begin{table*}
\vspace{-5mm}
\centering
\scriptsize
\begin{tabular}{crcccccc} 
\toprule
Sce.         & Metrics   & Random       & APF          & WMA-RRT      & Voronoi.    & {\name}          \\ 
\midrule
\multirow{3}{*}{\tabincell{c}{Training\\Sce.}} 
                                               & \emph{Mut. Over.} $\downarrow$ & 0.66(0.01)  & 0.61(0.01)   & 0.61(0.01)   & 0.44(0.01)          & \textbf{0.42(0.01)}     \\ 
\cmidrule{2-7}
                                               & \emph{Steps} $\downarrow$      & 273.56(1.38) & 251.41(3.15) & 268.20(2.24) & 237.04(2.95)        & \textbf{158.55(2.25)}   \\ 
\cmidrule{2-7}
                                               & \emph{Coverage} $\uparrow$     & 0.86(0.00)   & 0.90(0.01)   & 0.87(0.01)   & 0.93(0.00)          & \textbf{0.97(0.00)}     \\ 
\midrule
\multirow{3}{*}{\tabincell{c}{Testing\\Sce.}} 
                                               & \emph{Mut. Over.} $\downarrow$ & 0.66(0.02) &   0.57(0.01) &   0.64(0.01) &   \textbf{0.51(0.01)} &         0.54(0.02)              \\ 
\cmidrule{2-7}
                                               & \emph{Steps} $\downarrow$      & 193.83(2.80) & 181.18(4.17) & 198.92(3.83) & 156.68(3.21) &       \textbf{144.16(2.52)}   \\ 
\cmidrule{2-7}
                                               & \emph{Coverage} $\uparrow$     & 0.93(0.00) &   0.93(0.01) &   0.91(0.01) &   \textbf{0.96(0.01) }& \textbf{0.96(0.00)}    \\
\bottomrule
\end{tabular}
\caption{Performance of {\name} and \emph{multi-agent baselines} with a fixed size of $N=2$ agents on both training and testing scenes.}
\label{tab: training_agents_multi}
\vspace{-5mm}
\end{table*}
\begin{table*}
	\centering
	\vspace{-5mm}
	\scriptsize
	\begin{tabular}{crccccc} 
		\toprule
		\# Agent & Metrics &  Random & Nearest & Utility  & RRT & {\name} \\ 
		\midrule
		\multicolumn{7}{c}{\textbf{Training \quad Scenes}} \\ 
		\midrule
		\multirow{3}{*}{3}                                 
		& \emph{Mut. Over.} $\downarrow$ & 0.54(0.01)    & 0.46(0.01)   & 0.58(0.00)   & 0.44(0.01)          & \textbf{0.42(0.01)}              \\ 
		\cmidrule{2-7}
		& \emph{Steps} $\downarrow$      & 221.29(1.80) & 188.58(2.02) & 180.82(2.25) & 155.13(3.26)        & \textbf{127.88(1.91) }  \\ 
		\cmidrule{2-7}
		& \emph{Coverage} $\uparrow$     & 0.82(0.01)   & 0.91(0.00)   & 0.94(0.00)   & 0.95(0.01)          & \textbf{0.97(0.00)}     \\ 
		\midrule
		\multirow{3}{*}{4}                                 
		& \emph{Mut. Over.} $\downarrow$ & 0.49(0.01)  & 0.43(0.01)   & 0.52(0.01)   & \textbf{0.36(0.01) }         & 0.42(0.01)              \\ 
		\cmidrule{2-7}
		& \emph{Steps} $\downarrow$      & 163.11(0.77) & 154.75(2.16) & 151.30(3.03) & 140.57(1.78)        & \textbf{114.75(1.69)}   \\ 
		\cmidrule{2-7}
		& \emph{Coverage} $\uparrow$     & 0.87(0.00)    & 0.88(0.01)   & 0.91(0.01)   & 0.92(0.01)          & \textbf{0.96(0.00)}     \\ 
		\midrule
		\multicolumn{7}{c}{\textbf{Testing \quad Scenes}} \\ 
		\midrule
		\multirow{3}{*}{3}                                           
		& \emph{Mut. Over.} $\downarrow$ & 0.55(0.01) &   0.51(0.01) &   0.59(0.01) &   \textbf{0.45(0.01)} &         0.53(0.01)     \\ 
		\cmidrule{2-7}
		& \emph{Steps} $\downarrow$      & 145.90(2.80) & 131.04(3.53) & 128.46(3.04) & 128.33(1.66) &       \textbf{122.48(2.22)}  \\ 
		\cmidrule{2-7}
		& \emph{Coverage} $\uparrow$     & 0.94(0.00) &   0.95(0.00) &   \textbf{0.96(0.00)} &   0.95(0.01) &         \textbf{0.96(0.00)}  \\ 
		\midrule
		\multirow{3}{*}{\begin{tabular}[c]{@{}c@{}}\\4\end{tabular}} 
		& \emph{Mut. Over.} $\downarrow$ & 0.48(0.01) &   0.46(0.01) &   0.54(0.01) &   \textbf{0.41(0.01)} &         0.50(0.01)    \\ 
		\cmidrule{2-7}
		& \emph{Steps} $\downarrow$      & 116.94(1.61) & \textbf{108.23(1.24)} & 110.14(1.93) & 111.30(1.58) &       109.07(2.02) \\ 
		\cmidrule{2-7}
		& \emph{Coverage} $\uparrow$     &  0.93(0.01) &   \textbf{0.94(0.00)} &   \textbf{0.94(0.01) }&   0.93(0.01) &         \textbf{0.94(0.00)  }  \\ 
		\bottomrule
	\end{tabular}
	\caption{Zero-shot generalization performance of {\name} trained with a fixed team size $N=2$ and \emph{single-agent methods} to novel team sizes on training and testing scenes.}
	\label{tab: unseen_agents_single}
\end{table*}
\begin{table*}
	\centering
	\vspace{-5mm}
	\scriptsize
	\begin{tabular}{crcccccc} 
		\toprule
		\# Agent & Metrics & Random       & APF          & WMA-RRT      & Voronoi  & {\name} \\ 
		\midrule
		\multicolumn{7}{c}{\textbf{Training \quad Scenes}} \\ 
		\midrule
		\multirow{3}{*}{3}                                 
		& \emph{Mut. Over.} $\downarrow$ & 0.54(0.01)   & 0.45(0.01)   & 0.54(0.01)   & \textbf{0.37(0.01)} & 0.42(0.01)              \\ 
		\cmidrule{2-7}
		& \emph{Steps} $\downarrow$      & 221.29(1.80) & 207.20(2.41) & 210.01(2.68) & 180.27(2.51)        & \textbf{127.88(1.91) }  \\ 
		\cmidrule{2-7}
		& \emph{Coverage} $\uparrow$     & 0.82(0.01)   & 0.87(0.01)   & 0.87(0.01)   & 0.95(0.00)          & \textbf{0.97(0.00)}     \\ 
		\midrule
		\multirow{3}{*}{4}                                 
		& \emph{Mut. Over.} $\downarrow$ & 0.49(0.01)   & 0.35(0.01)   & 0.49(0.01)   & \textbf{0.34(0.01)} & 0.42(0.01)              \\ 
		\cmidrule{2-7}
		& \emph{Steps} $\downarrow$      & 163.11(0.77) & 170.59(1.06) & 168.07(1.41) & 147.01(2.38)        & \textbf{114.75(1.69)}   \\ 
		\cmidrule{2-7}
		& \emph{Coverage} $\uparrow$     & 0.87(0.00)   & 0.79(0.01)   & 0.82(0.01)   & 0.93(0.00)          & \textbf{0.96(0.00)}     \\ 
		\midrule
		\multicolumn{7}{c}{\textbf{Testing \quad Scenes}} \\ 
		\midrule
		\multirow{3}{*}{3}                                           
		& \emph{Mut. Over.} $\downarrow$ &  0.55(0.01) &   \textbf{0.40(0.01)} &   0.56(0.01) &   0.43(0.01) &         0.53(0.01)    \\ 
		\cmidrule{2-7}
		& \emph{Steps} $\downarrow$      & 145.90(2.80) & 152.62(3.71) & 161.59(3.60) & \textbf{119.98(2.31) }&       122.48(2.22) \\ 
		\cmidrule{2-7}
		& \emph{Coverage} $\uparrow$     & 0.94(0.00) &   0.92(0.01) &   0.89(0.02) &   \textbf{0.96(0.00)} &  \textbf{  0.96(0.00)}    \\ 
		\midrule
		\multirow{3}{*}{\begin{tabular}[c]{@{}c@{}}\\4\end{tabular}} 
		& \emph{Mut. Over.} $\downarrow$ & 0.48(0.01) &   \textbf{0.30(0.01)} &   0.52(0.01) &   0.39(0.01) &         0.50(0.01)    \\ 
		\cmidrule{2-7}
		& \emph{Steps} $\downarrow$      & 116.94(1.61) & 133.68(1.35) & 136.88(3.08) & \textbf{101.90(2.36) }&       109.07(2.02)  \\ 
		\cmidrule{2-7}
		& \emph{Coverage} $\uparrow$     & 0.93(0.01) &   0.88(0.01) &   0.84(0.02) &   \textbf{0.95(0.00) }&         0.94(0.00)    \\ 
		\bottomrule
	\end{tabular}
	\caption{Zero-shot generalization performance of {\name} trained with a fixed team size $N=2$ and \emph{multi-agent methods} to novel team sizes on training and testing scenes.}
	\label{tab: unseen_agents_multi}
\end{table*}
\vspace{-1mm}
\subsubsection{Trained with Team size = 3}
Here we additionally report the performance of all the baseline methods and {\name} trained with a fixed team size of $N=3$ agents on 9 representative training scenes and testing scenes in~\cref{tab: training_3agents_single} and \cref{tab: training_3agents_multi}. Except for comparable performance to Voronoi on testing scenes, {\name} consistently outperforms other planning-based baselines in all metrics on both training scenes and testing scenes.
When training with team size $3$, {\name} also exhibits surprising zero-shot transfer ability to various team sizes on training and testing scenes as shown in ~\cref{tab: unseen_3agents_single} and \cref{tab: unseen_3agents_multi}. More concretely, {\name} trained with a fixed team size $N=3$ shows much better performance than all planning-based methods on training scenes and comparable performance on testing scenes. 
Besides, we can observe that {\name} trained with a fixed team size $N=2$ and {\name} trained with a fixed team size $N=3$ are better than each other with corresponding training team size. {\name} trained with a fixed team size $N=3$ performs better in 4-agent case, showing $0.07$ less $Mutual Overlap$ and $8$ fewer $Steps$ on training scenes and $0.04$ less $Mutual Overlap$ and $6$ fewer $Steps$ on testing scenes.



\begin{table*}
\vspace{-12mm}
\centering
\scriptsize
\begin{tabular}{crccccc} 
\toprule
Sce.         & Metrics &   Random  & Nearest          & Utility      & RRT   & {\name}          \\ 
\midrule
\multirow{3}{*}{\tabincell{c}{Training\\Sce.}} 
                                                   & \emph{Mut. Over.} $\downarrow$ & 0.54\scriptsize{(0.01)}   & 0.46\scriptsize{(0.01)}   & 0.58\scriptsize{(0.00)}   & 0.44\scriptsize{(0.01)}          & \textbf{0.33\scriptsize{(0.01)}}    \\ 
\cmidrule{2-7}
                                                   & \emph{Steps} $\downarrow$      & 221.29\scriptsize{(1.80)} & 188.58\scriptsize{(2.02)} & 180.82\scriptsize{(2.25)} & 155.13\scriptsize{(3.26)}        & \textbf{121.99\scriptsize{(1.91)}}  \\ 
\cmidrule{2-7}
                                                   & \emph{Coverage} $\uparrow$     & 0.82\scriptsize{(0.01)}   & 0.91\scriptsize{(0.00)}   & 0.94\scriptsize{(0.00)}   & 0.95\scriptsize{(0.01)}          & \textbf{0.97\scriptsize{(0.00)}}    \\ 
\midrule
\multirow{3}{*}{\tabincell{c}{Testing\\Sce.}} 
                                                   & \emph{Mut. Over.} $\downarrow$ & 0.55(0.01) &   0.51(0.01) &   0.59(0.01) &   \textbf{0.45(0.01)} &         0.48(0.01)             \\ 
\cmidrule{2-7}
                                                   & \emph{Steps} $\downarrow$      & 145.90(2.80) & 131.04(3.53) & 128.46(3.04) & 128.33(1.66) &       \textbf{121.62(1.96)}  \\ 
\cmidrule{2-7}
                                                   & \emph{Coverage} $\uparrow$     & 0.94(0.00) &   0.95(0.00) &   \textbf{0.96(0.00)} &   0.95(0.01) &         \textbf{0.96(0.00)}  \\
\bottomrule
\end{tabular}
\caption{Performance of {\name} and \emph{single-agent baselines} with a fixed size of $N=3$ agents on both training and testing scenes.}
\label{tab: training_3agents_single}
\end{table*}
\begin{table*}
\vspace{-3mm}
\centering
\scriptsize
\begin{tabular}{crcccccc} 
\toprule
Sce.         & Metrics   & Random       & APF          & WMA-RRT      & Voronoi    & {\name}          \\ 
\midrule
\multirow{3}{*}{\tabincell{c}{Training\\Sce.}} 
                                               & \emph{Mut. Over.} $\downarrow$ & 0.54(0.01)   &  0.45(0.01)   & 0.54(0.01)   & 0.37(0.01)          & \textbf{0.33(0.01)}     \\ 
\cmidrule{2-7}
                                               & \emph{Steps} $\downarrow$      & 221.29(1.80) & 207.20(2.41) & 210.01(2.68)) & 180.27(2.51)        & \textbf{121.99(1.91)}   \\ 
\cmidrule{2-7}
                                               & \emph{Coverage} $\uparrow$     & 0.82(0.01)   & 0.87(0.01)   & 0.87(0.01)   & 0.95(0.00)          & \textbf{0.97(0.00)}     \\ 
\midrule
\multirow{3}{*}{\tabincell{c}{Testing\\Sce.}} 
                                               & \emph{Mut. Over.} $\downarrow$ & 0.55(0.01) &   0.40(0.01) &   0.56(0.01) &   \textbf{0.43(0.01)} &         0.48(0.01)              \\ 
\cmidrule{2-7}
                                               & \emph{Steps} $\downarrow$      & 145.90(2.80) & 152.62(3.71) & 161.59(3.60) & \textbf{119.98(2.31)} &       121.62(1.96)   \\ 
\cmidrule{2-7}
                                               & \emph{Coverage} $\uparrow$     & 0.94(0.00) &   0.92(0.01) &   0.89(0.02) &   \textbf{0.96(0.00)} &  \textbf{ 0.96(0.00)  }   \\
\bottomrule
\end{tabular}
\caption{Performance of {\name} and \emph{multi-agent baselines} with a fixed size of $N=3$ agents on both training and testing scenes. }
\label{tab: training_3agents_multi}
\vspace{-6mm}
\end{table*}
\begin{table*}
\centering
\vspace{-5mm}
\scriptsize
\begin{tabular}{crccccc} 
\toprule
\# Agent & Metrics &  Random & Nearest & Utility  & RRT & {\name} \\ 
\midrule
\multicolumn{7}{c}{\textbf{Training \quad Scenes}} \\ 
\midrule
\multirow{3}{*}{2}                                
                                                   & \emph{Mut. Over.} $\downarrow$ & 0.66(0.01)   & 0.53(0.02) &   0.68(0.01) &   0.53(0.02) &          \textbf{0.33(0.01)}              \\ 
\cmidrule{2-7}
                                                   & \emph{Steps} $\downarrow$      & 273.56(1.38) & 246.79(3.90) & 236.15(3.61) & 199.59(3.27)         & \textbf{167.24(2.12) }  \\ 
\cmidrule{2-7}
                                                   & \emph{Coverage} $\uparrow$     & 0.86(0.00)   & 0.91(0.01) &   0.92(0.01) &   0.96(0.00)& \textbf{0.96(0.00)}     \\ 
\midrule
\multirow{3}{*}{4}                                
                                                   & \emph{Mut. Over.} $\downarrow$ & 0.49(0.01) &   0.43(0.01) &   0.52(0.01) &   0.36(0.01) &   \textbf{0.34(0.01) }          \\ 
\cmidrule{2-7}
                                                   & \emph{Steps} $\downarrow$      & 163.11(0.77) & 154.75(2.16) & 151.30(3.03) & 140.57(1.78) & \textbf{106.12(2.19)}   \\ 
\cmidrule{2-7}
                                                   & \emph{Coverage} $\uparrow$     & 0.87(0.00) &   0.88(0.01) &   0.91(0.01) &   0.92(0.01) &   \textbf{0.96(0.00)}     \\ 
\midrule
\multicolumn{7}{c}{\textbf{Testing \quad Scenes}} \\ 
\midrule
\multirow{3}{*}{2}                                           
                                                             & \emph{Mut. Over.} $\downarrow$ &   0.66(0.02) &   0.58(0.01) &   0.69(0.01) &   0.57(0.02) &         \textbf{0.52(0.01)}   \\ 
\cmidrule{2-7}
                                                             & \emph{Steps} $\downarrow$      & 193.83(2.80) & 166.23(3.96) & 161.28(2.32) & 157.29(2.59) &       \textbf{154.95(2.95) }  \\ 
\cmidrule{2-7}
                                                             & \emph{Coverage} $\uparrow$     &  0.93(0.00) &   0.95(0.00) &   0.95(0.00) &   0.95(0.01) &         \textbf{0.96(0.00)}    \\ 
\midrule
\multirow{3}{*}{\begin{tabular}[c]{@{}c@{}}\\4\end{tabular}} 
                                                             & \emph{Mut. Over.} $\downarrow$ & 0.48(0.01) &   0.46(0.01) &   0.54(0.01) &   \textbf{0.41(0.01)} &         0.46(0.01)   \\ 
\cmidrule{2-7}
                                                             & \emph{Steps} $\downarrow$      & 116.94(1.61) & 108.23(1.24) & 110.14(1.93) & 111.30(1.58) &       \textbf{103.52(1.98)}  \\ 
\cmidrule{2-7}
                                                             & \emph{Coverage} $\uparrow$     & 0.93(0.01) &   0.94(0.00) &   0.94(0.01) &   0.93(0.01) &         \textbf{0.95(0.00)}    \\ 
\bottomrule
\end{tabular}
\caption{Zero-shot generalization performance of {\name} trained with a fixed team size $N=3$ and \emph{single-agent methods} to novel team sizes on training and testing scenes. }
\label{tab: unseen_3agents_single}
\end{table*}
\begin{table*}
\centering
\vspace{-5mm}
\scriptsize
\begin{tabular}{crcccccc} 
\toprule
\# Agent & Metrics & Random       & APF          & WMA-RRT      & Voronoi  & {\name} \\ 
\midrule
\multicolumn{7}{c}{\textbf{Training \quad Scenes}} \\ 
\midrule
\multirow{3}{*}{2}                                 
                                                   & \emph{Mut. Over.} $\downarrow$ & 0.66(0.01)   & 0.61(0.01)   & 0.61(0.01)   &  0.44(0.01) & \textbf{0.33(0.01)}              \\ 
\cmidrule{2-7}
                                                   & \emph{Steps} $\downarrow$      &273.56(1.38) & 251.41(3.15) & 268.20(2.24) & 237.04(2.95)        & \textbf{ 167.24(2.12) }  \\ 
\cmidrule{2-7}
                                                   & \emph{Coverage} $\uparrow$     & 0.86(0.00)   & 0.90(0.01)   & 0.87(0.01)   & 0.93(0.00)          & \textbf{0.96(0.00)}     \\ 
\midrule
\multirow{3}{*}{4}                                
                                                   & \emph{Mut. Over.} $\downarrow$ & 0.49(0.01) &   0.35(0.01) &   0.49(0.01) &   \textbf{0.34(0.01)} &   \textbf{0.34(0.01)}             \\ 
\cmidrule{2-7}
                                                   & \emph{Steps} $\downarrow$      & 163.11(0.77) & 170.59(1.06) & 168.07(1.41) & 147.01(2.38) & \textbf{106.12(2.19)}   \\ 
\cmidrule{2-7}
                                                   & \emph{Coverage} $\uparrow$     & 0.87(0.00) &   0.79(0.01) &   0.82(0.01) &   0.93(0.00) &   \textbf{0.96(0.00)}     \\ 
\midrule
\multicolumn{7}{c}{\textbf{Testing \quad Scenes}} \\ 
\midrule
\multirow{3}{*}{2}                                           
                                                             & \emph{Mut. Over.} $\downarrow$ &  0.66(0.02) &   0.57(0.01) &   0.64(0.01) &   \textbf{0.51(0.02)} &         0.52(0.01)    \\ 
\cmidrule{2-7}
                                                             & \emph{Steps} $\downarrow$      & 193.83(2.80) & 181.18(4.17) & 198.92(3.83) & 156.68(3.21) &       \textbf{154.95(2.95)} \\ 
\cmidrule{2-7}
                                                             & \emph{Coverage} $\uparrow$     & 0.93(0.00) &   0.93(0.01) &   0.91(0.01) &   \textbf{0.96(0.01)} &         \textbf{0.96(0.00)}    \\ 
\midrule
\multirow{3}{*}{\begin{tabular}[c]{@{}c@{}}\\4\end{tabular}} 
                                                             & \emph{Mut. Over.} $\downarrow$ & 0.48(0.01) &   \textbf{0.30(0.01)} &   0.52(0.01) &   0.39(0.01) &         0.46(0.01)   \\ 
\cmidrule{2-7}
                                                             & \emph{Steps} $\downarrow$      & 116.94(1.61) & 133.68(1.35) & 136.88(3.08) & \textbf{101.90(2.36)} &       103.52(1.98)  \\ 
\cmidrule{2-7}
                                                             & \emph{Coverage} $\uparrow$     & 0.93(0.01) &   0.88(0.01) &   0.84(0.02) &   \textbf{0.95(0.00)} &         \textbf{0.95(0.00)}   \\ 
\bottomrule
\end{tabular}
\caption{Zero-shot generalization performance of {\name} trained with a fixed team size $N=3$ and \emph{multi-agent methods} to novel team sizes on training and testing scenes. }
\label{tab: unseen_3agents_multi}
\end{table*}






\vspace{-2mm}
\subsection{Varying Team Size}
\vspace{-1mm}

We further consider the setting where the team size varies within an episode. We summarize the zero-shot generalization performance of {\name} compared with the planning-based baselines on training scenes in~\cref{tab: training_maps_vary_single} and \cref{tab: training_maps_vary_multi}. We use "$N_1\Rightarrow N_2$" to denote that each episode starts with $N_1$ agents and the team size immediately switches to $N_2$ after $90$ timesteps. More concretely, in an episode, max($N_1$, $N_2$) agents are set in the beginning, and $N_2$-$N_1$ agents are unable to move until timesteps 90 reaches in the increased team size scenarios. The setting is reversed in the decreased ones. We remark that {\name} trained by fixed team size $N=2$ or $N=3$ is separately presented in experiments, which is called {\name}{\scriptsize{(N=2)}} or {\name}{\scriptsize{(N=3)}}.

In scenarios where the team size increases, though RRT still performs the best among the planning-based baselines, {\name} outperforms RRT with a clear margin for $25$ fewer $Steps$ in $2 \Rightarrow 4$ setting and $35$ fewer $Steps$ in others. While as the team size decreases, the performance between {\name} and classical methods varies more widely, which shows that {\name} has a $35$ fewer $Steps$ in the comparison of the best baseline, RRT. Besides, {\name} has the best result in the metrics of mutual overlap ratio and coverage. We consider the case is more challenging where the team size decreases and the share information gain becomes less shapely, therefore the baselines could not adjust the strategy immediately.

\begin{table}
\centering
\scriptsize
\vspace{-12mm}
\resizebox{1.0\textwidth}{!}{
\begin{tabular}{crcccccc} 
\toprule
\# Agent                 & Metrics & Random       & Nearest          & Utility      & RRT          & \tabincell{c}{{\name}\\(N=2)}    &   \tabincell{c}{{\name}\\(N=3)} \\ 
\midrule
\multicolumn{8}{c}{\textbf{Increase Number of Agents}}                                                                                 \\ 
\midrule
\multirow{3}{*}{$2 \Rightarrow 3$} 
                         & \emph{Mut. Over.} $\downarrow$ & 0.54(0.01) &   0.43(0.01) &   0.56(0.01) &   0.36(0.01) &   0.30(0.01) &   \textbf{0.25(0.01)} \\ 
\cmidrule{2-8}
                         & \emph{Steps} $\downarrow$  & 223.63(2.16) & 211.73(1.96) & 210.88(1.30) & 185.94(1.83) & 148.82(2.01) & \textbf{146.34(1.76)} \\ 
\cmidrule{2-8}
                         & \emph{Coverage} $\uparrow$ & 0.86(0.01) &   0.89(0.01) &   0.90(0.00) &   0.94(0.00) &   \textbf{0.96(0.00)} &   \textbf{0.96(0.00)} \\ 
\midrule
\multirow{3}{*}{$2 \Rightarrow 4$} 
                         & \emph{Mut. Over.} $\downarrow$ & 0.42(0.01) &   0.37(0.01) &   0.47(0.00) &   0.31(0.01) &   0.26(0.01) &   \textbf{0.22(0.01)} \\ 
\cmidrule{2-8}
                         & \emph{Steps} $\downarrow$  & 175.89(0.64) & 174.06(0.72) & 174.55(0.69) & 165.43(0.97) & 142.96(1.24) & \textbf{138.22(1.36)} \\ 
\cmidrule{2-8}
                         & \emph{Coverage} $\uparrow$  & 0.82(0.01) &   0.81(0.01) &   0.81(0.00) &   0.88(0.01) &   \textbf{0.94(0.00)} &   \textbf{0.94(0.00)} \\
\midrule
\multirow{4}{*}{$3 \Rightarrow 4$} 
                         & \emph{Mut. Over.} $\downarrow$ & 0.45(0.01) &   0.38(0.01) &   0.49(0.01) &   0.26(0.01) &   0.29(0.01) &   \textbf{0.24(0.01)} \\ 
\cmidrule{2-8}
                         & \emph{Steps} $\downarrow$ & 170.90(1.19) & 165.85(0.80) & 165.82(0.72) & 155.15(2.18) & 125.26(1.46) & \textbf{119.03(1.36)}   \\ 
\cmidrule{2-8}
                         & \emph{Coverage} $\uparrow$ & 0.85(0.01) &   0.85(0.00) &   0.87(0.01) &   0.90(0.01) &   0.95(0.00) &   \textbf{0.96(0.00)} \\
\midrule
\multicolumn{8}{c}{\textbf{Decrease Number of Agents}}                                                                                   \\ 
\midrule
\multirow{3}{*}{$3 \Rightarrow 2$}  
                         & \emph{Mut. Over.} $\downarrow$ & 0.48(0.01) &   0.39(0.01) &   0.48(0.01) &   0.35(0.01) &   0.41(0.01) &   \textbf{0.33(0.01)}\\ 
\cmidrule{2-8}
                         & \emph{Steps} $\downarrow$ & 225.51(2.57) & 213.16(1.65) & 209.87(1.75) & 187.93(1.98) & 145.14(2.83) & \textbf{142.09(1.98)} \\ 
\cmidrule{2-8}
                         & \emph{Coverage} $\uparrow$ & 0.84(0.01) &   0.86(0.01) &   0.88(0.00) &   0.91(0.00) &   \textbf{0.95(0.00)} &   \textbf{0.95(0.00)}  \\ 
\midrule
\multirow{3}{*}{$4 \Rightarrow 2$} 
                         & \emph{Mut. Over.} $\downarrow$ & 0.41(0.01) &   0.36(0.01) &   0.42(0.01) &   \textbf{0.32(0.01)} &   0.40(0.01) &   0.33(0.01)\\ 
\cmidrule{2-8}
                         & \emph{Steps} $\downarrow$  & 173.36(1.08) & 168.89(1.12) & 167.54(1.30) & 157.40(2.56) & 127.35(2.08) & \textbf{118.93(2.30)}\\ 
\cmidrule{2-8}
                         & \emph{Coverage} $\uparrow$ &  0.81(0.00) &   0.82(0.01) &   0.83(0.01) &   0.86(0.01) &   \textbf{0.93(0.00)} &   \textbf{0.93(0.00)} \\
\midrule
\multirow{3}{*}{$4 \Rightarrow 3$} 
                         & \emph{Mut. Over.} $\downarrow$ &  0.44(0.01) &   0.39(0.01) &   0.47(0.00) &   \textbf{0.33(0.01)} &   0.41(0.01) &   \textbf{0.33(0.01)}  \\ 
\cmidrule{2-8}
                        & \emph{Steps} $\downarrow$  & 168.32(1.24) & 161.87(1.08) & 159.96(1.48) & 147.61(1.78) & 119.59(2.31) & \textbf{111.24(1.54)} \\ 
\cmidrule{2-8}
                        & \emph{Coverage} $\uparrow$ &  0.85(0.00) &   0.85(0.00) &   0.88(0.01) &   0.90(0.01) &   \textbf{0.95(0.00)} &   \textbf{0.95(0.00)}\\
\bottomrule
\end{tabular}}
\caption{Performance of {\name} and \emph{single-agent baselines} with a varying team size on training scenes.}
\label{tab: training_maps_vary_single}
\vspace{-5mm}
\end{table}
\begin{table}
\centering
\scriptsize
\vspace{-8mm}
\resizebox{1.0\textwidth}{!}{
\begin{tabular}{crcccccc} 
\toprule
\# Agent                 & Metrics & Random       & APF          & WMA-RRT      & Voronoi          & \tabincell{c}{{\name}\\(N=2)}    &   \tabincell{c}{{\name}\\(N=3)}  \\ 
\midrule
\multicolumn{8}{c}{\textbf{Increase Number of Agents}}                                                                                 \\ 
\midrule
\multirow{3}{*}{$2 \Rightarrow 3$} 
                         & \emph{Mut. Over.} $\downarrow$ & 0.54(0.01) &   0.42(0.00) &   0.49(0.00) &   0.35(0.01) &   0.30(0.01) &   \textbf{0.25(0.01)} \\ 
\cmidrule{2-8}
                         & \emph{Steps} $\downarrow$  & 223.63(2.16) & 225.35(0.97) & 221.85(0.00) & 200.91(2.32) & 148.82(2.01) & \textbf{146.34(1.76)} \\ 
\cmidrule{2-8}
                         & \emph{Coverage} $\uparrow$ & 0.86(0.01) &   0.82(0.01) &   0.85(0.00) &   0.92(0.00) &   \textbf{0.96(0.00)} &   \textbf{0.96(0.00)} \\ 
\midrule
\multirow{3}{*}{$2 \Rightarrow 4$} 
                         & \emph{Mut. Over.} $\downarrow$ & 0.42(0.01) &   0.31(0.01) &   0.44(0.01) &   0.30(0.01) &   0.26(0.01) &   \textbf{0.22(0.01)} \\ 
\cmidrule{2-8}
                         & \emph{Steps} $\downarrow$  & 175.89(0.64) & 178.77(0.13) & 178.91(0.33) & 170.50(0.88) & 142.96(1.24) & \textbf{138.22(1.36)} \\ 
\cmidrule{2-8}
                         & \emph{Coverage} $\uparrow$  & 0.82(0.01) &   0.68(0.01) &   0.67(0.01) &   0.85(0.01) &   \textbf{0.94(0.00)} &   \textbf{0.94(0.00)} \\
\midrule
\multirow{3}{*}{$3 \Rightarrow 4$}
                         & \emph{Mut. Over.} $\downarrow$ & 0.45(0.01) &   0.32(0.01) &   0.45(0.00) &   0.31(0.01) &   0.29(0.01) &   \textbf{0.24(0.01)}\\ 
\cmidrule{2-8}
                         & \emph{Steps} $\downarrow$ & 170.90(1.19) & 175.35(0.56) & 173.64(0.41) & 159.23(1.50) & 125.26(1.46) &  \textbf{119.03(1.36)}  \\ 
\cmidrule{2-8}
                         & \emph{Coverage} $\uparrow$ & 0.85(0.01) &   0.74(0.01) &   0.79(0.00) &   0.90(0.00) &   0.95(0.00) &   \textbf{0.96(0.00)} \\
\midrule
\multicolumn{8}{c}{\textbf{Decrease Number of Agents}}                                                                                   \\ 
\midrule
\multirow{3}{*}{$3 \Rightarrow 2$}  
                         & \emph{Mut. Over.} $\downarrow$ & 0.48(0.01) &   0.38(0.01) &   0.44(0.00) &   \textbf{0.33(0.01)} &   0.41(0.01) &   \textbf{0.33(0.01)} \\ 
\cmidrule{2-8}
                         & \emph{Steps} $\downarrow$ & 225.51(2.57) & 225.51(1.32) & 226.14(0.00) & 206.94(2.50) & 145.14(2.83) & \textbf{142.09(1.98)}  \\ 
\cmidrule{2-8}
                         & \emph{Coverage} $\uparrow$ & 0.84(0.01) &   0.78(0.01) &   0.81(0.00) &   0.89(0.01) &   \textbf{0.95(0.00)} &   \textbf{0.95(0.00)} \\ 
\midrule
\multirow{3}{*}{$4 \Rightarrow 2$} 
                         & \emph{Mut. Over.} $\downarrow$ & 0.41(0.01) &   0.32(0.01) &   0.40(0.00) &   \textbf{0.30(0.01)} &   0.40(0.01) &   0.33(0.01)\\ 
\cmidrule{2-8}
                         & \emph{Steps} $\downarrow$  & 173.36(1.08) & 176.79(0.65) & 176.38(0.00) & 165.23(2.80) & 127.35(2.08) & \textbf{118.93(2.30)}\\ 
\cmidrule{2-8}
                         & \emph{Coverage} $\uparrow$ &  0.81(0.00) &   0.68(0.01) &   0.73(0.00) &   0.83(0.01) &   \textbf{0.93(0.00)} &   \textbf{0.93(0.00)} \\
\midrule
\multirow{3}{*}{$4 \Rightarrow 3$} 
                         & \emph{Mut. Over.} $\downarrow$ &  0.44(0.01) &   0.33(0.01) &   0.44(0.00) &   \textbf{0.32(0.01)} &   0.41(0.01) &   0.33(0.01)  \\ 
\cmidrule{2-8}
                        & \emph{Steps} $\downarrow$  & 168.32(1.24) & 173.94(0.81) & 171.85(0.00) & 155.65(2.79) & 119.59(2.31) & \textbf{111.24(1.54)} \\ 
\cmidrule{2-8}
                        & \emph{Coverage} $\uparrow$ &  0.85(0.00) &   0.73(0.01) &   0.78(0.00) &   0.89(0.01) &   \textbf{0.95(0.00)} &   \textbf{0.95(0.00)}\\
\bottomrule
\end{tabular}}
\caption{Performance of {\name} and \emph{multi-agent baselines} with a varying team size on training scenes.}
\label{tab: training_maps_vary_multi}
\vspace{-5mm}
\end{table}

When we compare {\name}{\scriptsize{(N=2)}} with {\name}{\scriptsize{(N=3)}}, {\name}{\scriptsize{(N=3)}} has a comparable performance with a lower $Mutual\ Overlap$ ratio to the other. It indicates that training with a fixed team size $N=3$ helps {\name} grasp the capability of cooperation better so that the strategy is more stable and inflexible in a varying team size situation. 

\begin{table}
\centering
\scriptsize
\vspace{-5mm}
\resizebox{1.0\textwidth}{!}{
\begin{tabular}{crcccccc} 
\toprule
\# Agent                 & Metrics & Random       & Nearest          & Utility      & RRT          & \tabincell{c}{{\name}\\(N=2)}    &   \tabincell{c}{{\name}\\(N=3)} \\ 
\midrule
\multicolumn{8}{c}{\textbf{Increase Number of Agents}}                                                                                 \\ 
\midrule
\multirow{3}{*}{$2 \Rightarrow 3$} 
                         & \emph{Mut. Over.} $\downarrow$ & 0.52(0.01) &   0.47(0.01) &   0.55(0.02) &   0.43(0.01) &         0.46(0.01)&\textbf{0.41(0.01)} \\ 
\cmidrule{2-8}
                         & \emph{Steps} $\downarrow$  &159.85(3.60) & 144.60(3.45) & 143.14(1.93) & 136.42(2.41) &       134.11(2.88)&\textbf{131.96(1.80)} \\ 
\cmidrule{2-8}
                         & \emph{Coverage} $\uparrow$ & 0.93(0.01) &   0.95(0.01) &   0.95(0.00) &   \textbf{0.96(0.01)} &         \textbf{0.96(0.00)}&\textbf{0.96(0.00)} \\ 
\midrule
\multirow{3}{*}{$2 \Rightarrow 4$} 
                         & \emph{Mut. Over.} $\downarrow$ & 0.43(0.01) &   0.40(0.01) &   0.47(0.01) &   0.38(0.01) &         0.43(0.01) &\textbf{0.37(0.01) }\\ 
\cmidrule{2-8}
                         & \emph{Steps} $\downarrow$  & 134.57(0.63) & 129.13(1.86) & 126.92(1.67) & 122.42(1.85) &       122.09(1.99)&\textbf{116.66(4.51)} \\ 
\cmidrule{2-8}
                         & \emph{Coverage} $\uparrow$  &  0.90(0.00) &   0.91(0.01) &   0.92(0.01) &   0.92(0.00) &         \textbf{0.93(0.01)}&\textbf{0.93(0.01)} \\
\midrule
\multirow{3}{*}{$3 \Rightarrow 4$} 
                         & \emph{Mut. Over.} $\downarrow$ &  0.46(0.01) &   0.42(0.01) &   0.49(0.01) &  \textbf{ 0.37(0.01) }&         0.45(0.01)& 0.39(0.00) \\ 
\cmidrule{2-8}
                         & \emph{Steps} $\downarrow$ & 125.08(1.35) & 116.96(1.42) & 115.44(3.38) & 119.02(1.32) &       114.84(1.56)&\textbf{110.04(0.54) } \\ 
\cmidrule{2-8}
                         & \emph{Coverage} $\uparrow$ &0.92(0.01) &   0.93(0.00) &   0.93(0.01) &   0.92(0.01) &         \textbf{0.94(0.00)}&\textbf{0.94(0.00)} \\
\midrule
\multicolumn{8}{c}{\textbf{Decrease Number of Agents}}                                                                                   \\ 
\midrule
\multirow{3}{*}{$3 \Rightarrow 2$}  
                         & \emph{Mut. Over.} $\downarrow$ &0.45(0.01) &   0.41(0.01) &   0.46(0.01) &   \textbf{0.39(0.01)} &         0.43(0.01)&0.40(0.00)\\ 
\cmidrule{2-8}
                         & \emph{Steps} $\downarrow$ & 167.98(2.06) & 149.41(2.59) & 146.73(3.44) & \textbf{139.52(3.74)} &       145.43(3.44) &146.93(2.93)\\ 
\cmidrule{2-8}
                         & \emph{Coverage} $\uparrow$ & 0.91(0.00) &   \textbf{0.94(0.00) }&   0.93(0.01) &   \textbf{0.94(0.01) }&         \textbf{0.94(0.01)}& \textbf{0.94(0.00)} \\ 
\midrule
\multirow{3}{*}{$4 \Rightarrow 2$} 
                         & \emph{Mut. Over.} $\downarrow$ & 0.36(0.01) &   0.33(0.01) &   0.37(0.01) &  \textbf{ 0.31(0.00)} &         0.37(0.01) &0.34(0.00) \\ 
\cmidrule{2-8}
                         & \emph{Steps} $\downarrow$  &140.37(1.80) & 128.11(1.72) & 127.40(0.49) & \textbf{125.84(0.83) }&       129.73(2.73)&127.30(1.55)\\ 
\cmidrule{2-8}
                         & \emph{Coverage} $\uparrow$ &  0.88(0.01) &   \textbf{0.90(0.01)} &  \textbf{ 0.90(0.01)} &   \textbf{0.90(0.00)} &         \textbf{0.90(0.01)}&\textbf{0.90(0.00)} \\
\midrule
\multirow{4}{*}{$4 \Rightarrow 3$} 
                         & \emph{Mut. Over.} $\downarrow$ &   0.43(0.01) &   0.39(0.00) &   0.44(0.01) &   \textbf{0.35(0.01) }&         0.43(0.01)&  0.39(0.01)\\ 
\cmidrule{2-8}
                        & \emph{Steps} $\downarrow$  &  127.46(1.64) & 117.44(0.89) & 114.56(2.28) & 119.70(1.88) &       116.74(1.38)& \textbf{112.72(2.13) }\\ 
\cmidrule{2-8}
                        & \emph{Coverage} $\uparrow$ &0.91(0.01) &   \textbf{0.93(0.00)} &   \textbf{0.93(0.01) }&   0.91(0.00) &         \textbf{0.93(0.01)}&\textbf{0.93(0.00)}\\
\bottomrule
\end{tabular}}
\caption{Performance of {\name} and \emph{single-agent baselines} with a varying team size on testing scenes.}
\label{tab: unseen_maps_vary_single}
\vspace{-5mm}
\end{table}
\begin{table}
\centering
\scriptsize
\vspace{-5mm}
\resizebox{1.0\textwidth}{!}{
\begin{tabular}{crcccccc} 
\toprule
\# Agent                 & Metrics & Random       & APF          & WMA-RRT      & Voronoi          & \tabincell{c}{{\name}\\(N=2)}    &   \tabincell{c}{{\name}\\(N=3)}  \\ 
\midrule
\multicolumn{8}{c}{\textbf{Increase Number of Agents}}                                                                                 \\ 
\midrule
\multirow{3}{*}{$2 \Rightarrow 3$} 
                         & \emph{Mut. Over.} $\downarrow$ &  0.52(0.01) &   0.38(0.02) &   0.44(0.01) &   \textbf{0.32(0.02) }&         0.46(0.01) &         0.41(0.01) \\ 
\cmidrule{2-8}
                         & \emph{Steps} $\downarrow$  & 159.85(3.60) & 167.87(2.85) & 176.89(5.70) & 148.12(6.69) &       134.11(2.88) &     \textbf{  131.96(1.80)} \\ 
\cmidrule{2-8}
                         & \emph{Coverage} $\uparrow$ & 0.93(0.01) &   0.89(0.01) &   0.88(0.01) &   0.92(0.05) &         \textbf{0.96(0.00)} &        \textbf{ 0.96(0.00)} \\ 
\midrule
\multirow{3}{*}{$2 \Rightarrow 4$} 
                         & \emph{Mut. Over.} $\downarrow$ &  0.43(0.01) &   0.26(0.01) &   0.33(0.01) &   \textbf{0.24(0.01)} &         0.43(0.01) &         0.37(0.01)\\ 
\cmidrule{2-8}
                         & \emph{Steps} $\downarrow$  & 134.57(0.63) & 148.48(1.01) & 157.48(1.50) & 130.49(1.81) &       122.09(1.99) &     \textbf{  116.66(4.51) }\\ 
\cmidrule{2-8}
                         & \emph{Coverage} $\uparrow$  &0.90(0.00) &   0.78(0.01) &   0.69(0.02) &   0.90(0.01) &         \textbf{0.93(0.01) }&        \textbf{ 0.93(0.01)} \\
\midrule
\multirow{3}{*}{$3 \Rightarrow 4$} 
                         & \emph{Mut. Over.} $\downarrow$ &  0.46(0.01) &   \textbf{0.28(0.01)} &   0.40(0.00) &   0.30(0.00) &         0.45(0.01) &         0.39(0.00)\\ 
\cmidrule{2-8}
                         & \emph{Steps} $\downarrow$ &125.08(1.35) & 139.82(1.77) & 141.80(1.53) & 114.99(1.04) &       114.84(1.56) &       \textbf{110.04(0.54) } \\ 
\cmidrule{2-8}
                         & \emph{Coverage} $\uparrow$ & 0.92(0.01) &   0.85(0.01) &   0.83(0.01) &  \textbf{ 0.94(0.00)} &       \textbf{  0.94(0.00) }&         \textbf{0.94(0.00)} \\
\midrule
\multicolumn{8}{c}{\textbf{Decrease Number of Agents}}                                                                                   \\ 
\midrule
\multirow{3}{*}{$3 \Rightarrow 2$}  
                         & \emph{Mut. Over.} $\downarrow$ & 0.45(0.01) &   \textbf{0.33(0.01)} &   0.51(0.01) &   0.42(0.01) &         0.43(0.01) &         0.40(0.00)\\ 
\cmidrule{2-8}
                         & \emph{Steps} $\downarrow$ & 167.98(2.06) & 178.43(2.25) & 171.87(2.53) & \textbf{133.77(2.83) }&       145.43(3.44) &       146.93(2.93)  \\ 
\cmidrule{2-8}
                         & \emph{Coverage} $\uparrow$ &  0.91(0.00) &   0.86(0.01) &   0.87(0.01) &  \textbf{ 0.95(0.01)} &         0.94(0.01) &         0.94(0.00)  \\ 
\midrule
\multirow{3}{*}{$4 \Rightarrow 2$} 
                         & \emph{Mut. Over.} $\downarrow$ &  0.36(0.01) &   \textbf{0.23(0.00)} &   0.47(0.01) &   0.38(0.00) &         0.37(0.01) &         0.34(0.00)\\ 
\cmidrule{2-8}
                         & \emph{Steps} $\downarrow$  & 140.37(1.80) & 151.69(1.60) & 144.66(1.46) &\textbf{ 115.97(1.93)} &       129.73(2.73) &       127.30(1.55)\\ 
\cmidrule{2-8}
                         & \emph{Coverage} $\uparrow$ &   0.88(0.01) &   0.78(0.01) &   0.81(0.01) &   \textbf{0.93(0.00)} &         0.90(0.01) &         0.90(0.00) \\
\midrule
\multirow{3}{*}{$4 \Rightarrow 3$} 
                         & \emph{Mut. Over.} $\downarrow$ &   0.43(0.01) &   \textbf{0.27(0.01)} &   0.48(0.00) &   0.39(0.01) &         0.43(0.01) &         0.39(0.01)  \\ 
\cmidrule{2-8}
                        & \emph{Steps} $\downarrow$  & 127.46(1.64) & 142.86(2.31) & 140.68(0.17) & \textbf{106.92(2.78)} &       116.74(1.38) &       112.72(2.13) \\ 
\cmidrule{2-8}
                        & \emph{Coverage} $\uparrow$ &  0.91(0.01) &   0.84(0.01) &   0.82(0.00) &  \textbf{ 0.94(0.01) }&         0.93(0.01) &         0.93(0.00)\\
\bottomrule
\end{tabular}}
\caption{Performance of {\name} and \emph{multi-agent baselines} with a varying team size on testing scenes. }
\label{tab: unseen_maps_vary_multi}
\vspace{-5mm}
\end{table}

We summarize the zero-shot generalization performance of {\name} compared with the planning-based baselines on testing scenes in~\cref{tab: unseen_maps_vary_single} and \cref{tab: unseen_maps_vary_multi}. In cases where the team size increases, {\name} still produces substantially better performances, especially $Steps$, which suggests that {\name} has the capability to adaptively adjust its strategy. Regarding the cases where the team size decreases, {\name} shows slightly worse performance than the best single-agent baseline, RRT, and the best multi-agent baseline, Voronoi. We remark that decreasing the team size is particularly challenging since the absence of some agents might immediately leave a large part of the house unexplored and consequently, the team should immediately update their original plan with drastically different goal assignments.

\newpage
\subsection{Case Studies}
\label{app: case}

We implemented several planning-based methods as baselines, including Nearest, Utility, APF, RRT and additionally a voronoi-based method plus a multi-agent variant of RRT, WMA-RRT. While these methods are respectively tested under environments with ideal assumptions, we empirically found some of them do not work well under our setting with realistic perception and noises.

\begin{figure}[ht]
	\centering
	\vspace{-2mm}
    \includegraphics[width=0.45\linewidth]{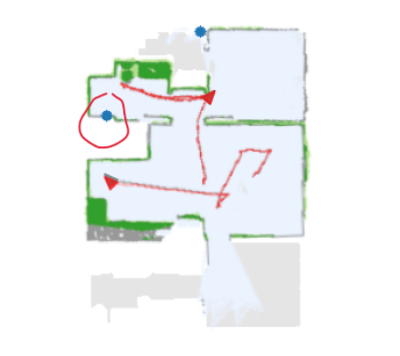}
    \vspace{-4mm}
	\centering 
	\caption{Sensitive to Realistic Noise. Utility, Nearest, APF and the voronoi-based method performs cell-level goal searching, the above picture demonstrates a case when the agent selects a cell that is estimated as "unexplored" because of mapping error.}
	\vspace{-4mm}
\label{fig:case1}
\end{figure}

\begin{figure}[ht]
	\centering
	\vspace{-2mm}
    \includegraphics[width=0.45\linewidth]{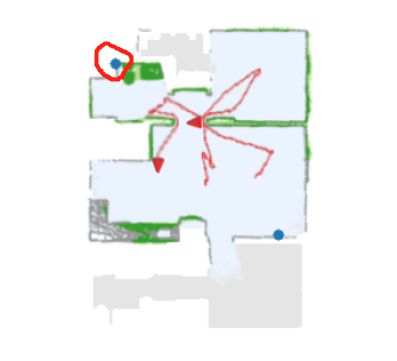}
    \vspace{-4mm}
	\centering 
	\caption{Faulty Estimation of Value. In the above situation, Utility selects a cell at the corner of the room as a global goal since it's nearby space is unexplored. However, with prior knowledge about building structures, one should infer that such point should not be chosen.}
	\vspace{-4mm}
\label{fig:case2}
\end{figure}

\begin{figure}[ht]
	\centering
	\vspace{-2mm}
    \includegraphics[width=0.45\linewidth]{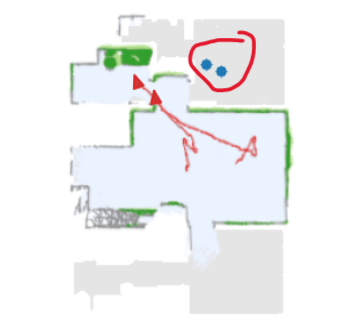}
    \vspace{-4mm}
	\centering 
	\caption{Poor Cooperation. RRT performs planning independently, it's possible that agents choose close frontiers as goals, leading to duplicated exploration.}
	\vspace{-4mm}
\label{fig:case3}
\end{figure}

\begin{figure}[ht]
	\centering
	\vspace{-6.5mm}
    \includegraphics[width=0.45\linewidth]{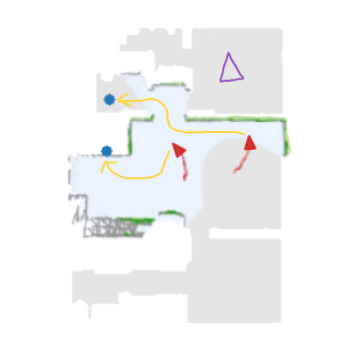}
    \vspace{-5.5mm}
	\centering 
	\caption{APF Resistance Force. APF achieves cooperation by introducing resistance force among agents. In above situation, the yellow lines demonstrate agents' paths to their corresponding global goal. Due to the resistance force, the agent on the left is forced to choose a sub-optimal global goal instead of frontiers in the more promising part indicated by the purple triangle.}
	\vspace{-4mm}
\label{fig:case4}
\end{figure}

\begin{figure}[ht]
	\centering
	\vspace{-1mm}
    \includegraphics[width=0.45\linewidth]{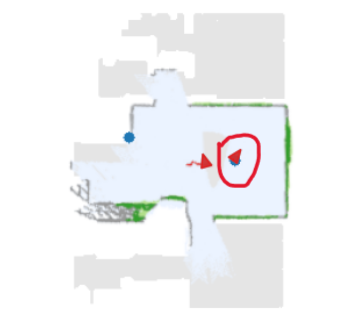}
    \vspace{-4mm}
	\centering 
	\caption{WMA-RRT Locking Mechanism. In WMA-RRT, agents cooperatively maintain a tree and use a locking mechanism to avoid duplicated exploration. The above picture shows a case where agent on the right has to stay where it is since the path is locked by another agent.}
	\vspace{-4mm}
\label{fig:case5}
\end{figure}

\begin{figure}[ht]
	\centering
	\vspace{-2mm}
    \includegraphics[width=0.45\linewidth]{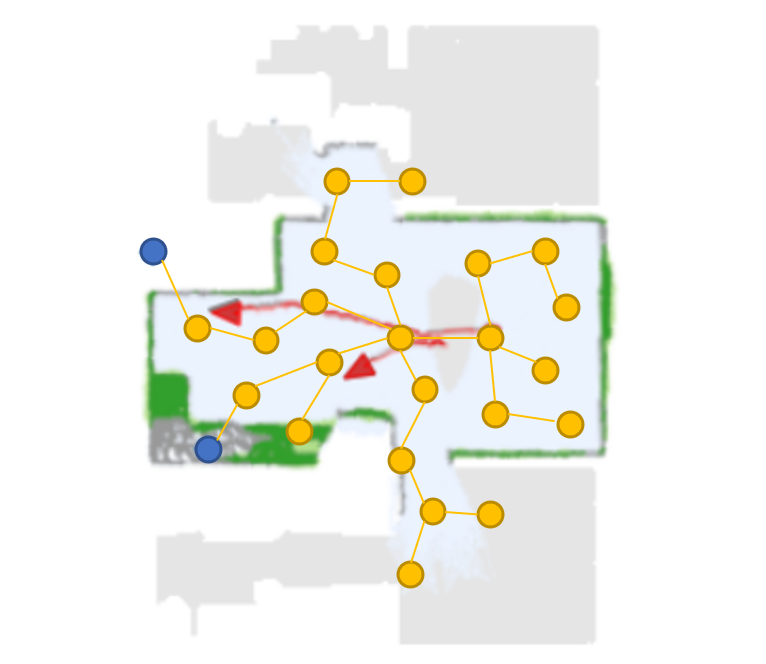}
    \vspace{-4mm}
	\centering 
	\caption{Incompatible with Active Mapping. Specially, WMA-RRT algorithm is inherently incompatible with the active mapping process. When new cells are classified as obstacles, some nodes and edges in the tree would become invalid and WMA-RRT could not adapt to such change. In the above picture, agents still try to approach selected points even the tree edges and nodes are no longer valid.}
	\vspace{-4mm}
\label{fig:case6}
\end{figure}

\begin{itemize}

\item \textbf{Sensitive to realistic noise.} Nearest, Utility, APF and the voronoi-based method all apply cell-level goal searching. We empirically found that they share a same failure mode, that is, trying to approach a cell that is wrongly estimated as explorable, even when the mapping only has minor cell-level error. The planning-based baseline RRT do not perform cell-level planning but in a geometric way, thus it’s robust to mapping noise.

\item\textbf{Naive estimation of future value.} Both Utility and RRT select candidate frontier with highest utility, which is computed as the number of unexplored cells within a small distance. Such estimation of future values is very short-sighted. In contrast, TANS, by using neural network, could perform more complicated inference about utility of exploring one part of the map.

\item \textbf{Strict restriction over robot behaviors.} Among the planning-based baselines, both APF and WMA-RRT adopt a formally-designed cooperation system. However their cooperation schemes both imposes great restriction to agents’ behaviors. APF includes resistance force among agents to avoid duplicated exploration. In WMA-RRT, agents share a same tree and follow a locking-and-searching method to do cooperation. However, in cases where it’s better for agents to go through the same place simultaneously, which is pretty common, the resistance force in APF and the locking mechanism in WMA-RRT prevent agents from the optimal strategy. Meanwhile, WMA-RRT restricts agents to follow paths in the shared tree, losing the benefit of accidental coverage brought by random search.

We provide case studies with corresponding graphics in Fig.~\ref{fig:case1}-\ref{fig:case6}. The red angles indicate the agents and their direction. The blue points indicate the global goals selected by the agents.

\end{itemize}

\newpage

\newpage

\newpage
\bibliography{iclr2022_conference}
\bibliographystyle{iclr2022_conference}

\end{document}